\documentclass[acmlarge,screen]{acmart}

\AtBeginDocument{%
  }

\setcopyright{none} 
\acmConference[]{}{}{} 
\acmBooktitle{} 
\acmPrice{} 
\acmDOI{} 
\acmISBN{} 
\makeatletter
\renewcommand\footnotetextcopyrightpermission[1]{} 
\@ACM@nonacmtrue
\renewcommand{\@authorsaddresses}{}
\renewcommand{\footnotetextauthorsaddresses}[1]{}
\makeatother

\usepackage{verbatim}
\begin{document}

\title{Evolution and The Knightian Blindspot of Machine Learning}

\author{Joel Lehman$^1$}
\authornote{Corresponding author: lehman.154@gmail.com; \href{https://www.second-nature.ai/}{Second Nature AI}$^1$, Cognizant AI Labs$^2$, Unaffiliated$^3$}
\author{Elliot Meyerson$^2$}
\author{Tarek El-Gaaly$^1$}
\author{Kenneth O. Stanley$^3$}
\author{Tarin Ziyaee$^1$}

\renewcommand{\shortauthors}{Lehman et al.}

%




\begin{abstract}
This paper claims that machine learning (ML) largely overlooks an important facet of
 general intelligence: robustness to a qualitatively unknown future in an open world.
Such robustness relates to the concept of Knightian uncertainty (KU) in 
economics, i.e.\ uncertainty that cannot be quantified, which is excluded from consideration in ML's key formalisms. This paper aims to identify this blind spot, argue its importance, and 
catalyze research into addressing it, which we believe is necessary to create truly robust open-world AI. 
To help illuminate the blind spot, we contrast 
 one area of ML, reinforcement learning (RL), with the process of biological evolution. 
Despite staggering ongoing progress, RL still struggles in open-world situations, often failing under unforeseen situations. For example, the idea of zero-shot transferring a self-driving car policy trained only in the US to the UK currently seems exceedingly ambitious. In dramatic contrast, biological evolution routinely produces agents that thrive within an open world, sometimes even to situations that are remarkably out-of-distribution (e.g.\ invasive species; or humans, who do undertake such zero-shot international driving). 
Interestingly, evolution achieves such robustness without explicit theory, formalisms, or mathematical gradients. We explore the assumptions underlying RL's typical formalisms, showing how they limit RL's engagement with the unknown unknowns characteristic of an ever-changing complex world.
Further, we identify mechanisms through which evolutionary processes foster robustness to novel and unpredictable challenges, and discuss potential pathways to algorithmically embody them, highlighting the promise of artificial life, open-endedness, and revisiting RL's core formalisms in service of managing Knightian uncertainty. The conclusion is that the intriguing remaining fragility of ML may result from blind spots induced by its formalisms, and that significant gains may result from direct confrontation with the challenge of KU.
\end{abstract}



\maketitle
\tableofcontents

\section{Introduction}

\begin{quote}
``Seeking an improvement that makes a difference in the shorter term, researchers seek to leverage their human knowledge of the domain, but the only thing that matters in the long run is the leveraging of computation. [...] We have to learn the bitter lesson that building in how we think we think does not work in the long run.'' \cite{sutton2019bitter}
\hfill --- Richard Sutton
\end{quote}

The above quote reflects the ascendancy in machine learning (ML) of methods that leverage increased computation through search and learning, a potentially ``bitter lesson'' for those tailoring ML algorithms through human domain knowledge \cite{sutton2019bitter}. For example, many sophisticated feature-construction methods for machine vision were made obsolete by deep learning on raw pixels \cite{zhang2014predicting,krizhevsky2012imagenet}, just as natural language processing was upended by the transformer revolution \cite{vaswani2017attention,brown2020language}. Yet unanswered by this bitter lesson is exactly \emph{how little} we should design by hand into ML systems. In other words, do there remain blind spots that exist within ML as a result of human thinking that could be overcome if yet more was given over to search \cite{clune2019ai}? If so, perhaps such blind spots could be remedied by taking the insights from the bitter lesson further (e.g.\ automating the search for architectures or learning algorithms), or illuminated by studying algorithms that already do so. This latter thread is pursued here.

In particular, this paper argues that there are subtle lessons still to digest from the algorithm that takes the bitter lesson to its logical conclusion: biological evolution, which imposed no prior domain knowledge, and instead effectively leveraged extreme amounts of computation to invent human intelligence on one leaf of its massively divergent search. The main idea is to highlight, through contrast with evolution, an intriguing blind spot in the formalisms of ML: the importance of robustness to an open-ended future. We hope to go beyond critique alone through suggesting mechanisms from nature, human intelligence, and threads of current ML research that may unlock future progress. 

Human engineering has surpassed biological evolution in many ways (e.g.\ the cargo capacity or speed of a jumbo jet, compared to birds). However our ingenuity has yet to match evolution in others, such as in engineering self-replicating machines, or, of central interest here: in matching the robustness of evolution's products. For example, in the late 1980s, zebra mussels made their way from Europe to the great lakes in the US, hitchhiking in the ballast tanks of ships \cite{ricciardi2003predicting}. They rapidly spread through the waterways, outcompeting native species and clogging water intake pipes. Such invasive species are not uncommon: Animals evolved for one environment, when placed into a significantly different one, can be successful, sometimes frustratingly so (to us) \cite{mooney2001evolutionary}. From an ML or robotics perspective, this level of out-of-distribution robustness is an amazing accomplishment; it is hard to imagine, for example, successfully zero-shot transferring a self-driving policy trained only on data from the US to the UK -- which us humans (another of evolution's products) regularly accomplish, albeit with visceral discomfort.

Like evolution, ML aspires towards agents that robustly function in \emph{open-world environments}, an important challenge spanning many critical applications, including social networks, chatbot assistants, home-assistance robots, self-driving cars, and robots acting in unstructured environments more broadly; indeed, competence within such open worlds is fundamental to ML's quest for artificial general intelligence (AGI) \cite{hughes2024open,parisi2019continual}. Yet despite their unprecedented pace of improvement in capabilities, ML systems still struggle in environments much different from those they were trained in: self-driving cars falter with rare situations that are mundane to humans (like understanding that traffic lights in the back of a trailer are not worrisome; or that a moving plastic bag is not a problem) \cite{siddiqui2022tesla,npr2024_tesla_probe,herger2024waymo,liu2024curse,hendrycks2021unsolved}; generative AI agents remain surprisingly fragile \cite{stewart2024surprisingly,zhao-etal-2024-improving,chen2024can,huang2023benchmarking,wang2024can,xie2024osworld,jimenez2023swe}; and chain-of-thought reasoning in 100 billion+ parameter language models frequently goes off the rails \cite{dhuliawala2023chain}. 
Others have remarked that this is a peculiar situation \cite{west2023generative,wu2023reasoning,vafa2024large}: An agent that is remarkably knowledgeable and capable of e.g.\ fluid philosophical discussion, can yet make rudimentary mistakes of generalization \cite{wu2023reasoning} or reliability \cite{huang2023survey,dziri2024faith}. One explanation to resolve this paradox is that facets of intelligence that are \emph{coupled} in humans may be \emph{decoupled} in machines: one or more such facets may still be missing from ML.

This paper thus aims to draw attention to an important property of intelligence largely ignored by ML: \emph{Robustness to a qualitatively unknown future in an open world}, which we relate to the idea of Knightian uncertainty (KU; \cite{sunstein2023knightian,knight1921risk,dimand2021keynes,keynes2013treatise,kay2020radical}), or roughly, \emph{unknown unknowns}. What emerges next in the real world (i.e.\ the next fashion, scientific idea, power outage, edge-case driving scenario, natural disaster, stock market boom or crash, etc.) is the vastly-complex product of billions of creative agents acting in an interconnected, complicated physical and digital world; there is much we reasonably cannot anticipate. Yet, the preponderance of work in robust ML aims to address \emph{known unknowns}, i.e.\ encouraging robustness to limited classes of experimenter-anticipated risks. For example, to apply techniques from safe reinforcement learning \cite{garcia2015comprehensive} an experimenter generally must anticipate the ways an environment can vary, and specify the type of risk within such environments they wish the policy to optimize. 
Such work is obviously useful and important. However, one pillar of general intelligence is robustness to the novel situations and risks that continually arise in an open world -- can we truly claim to be approaching ``AGI'' without such a capacity, which is central to the survival of animal species as well as human civilization?

We define an agent's robustness to KU as its capacity to succeed or at least fail gracefully in the face of novel unforeseeable situations. For example, the first zebra mussels that arrived from Europe in the great lakes faced situations far outside their window of direct evolutionary exposure, and yet thrived. In particular, this paper investigates assumptions in ML made for conceptual and mathematical convenience that limit robustness to plausible, if unlikely, future environmental disturbances. These assumptions are then contrasted with mechanisms from biological evolution that enable broader robustness. We focus centrally on reinforcement learning (RL), where the typical objective is to find a policy for an agent that maximizes expected reward, although many facets of this analysis apply also to the settings of unsupervised and supervised learning.

A central conclusion is that RL's engagement with the problem of KU is obscured by two main factors. First, thought in RL is often shaped by the field's formalisms, which tend to axiomatically rule out KU as a possibility; it is human to take for granted the water that immerses us \cite{abel2024three,kuhn1997structure}. For example, Markov decision processes (MDPs) and typical RL optimization objectives make simplifying closed-world assumptions: that the deployment environment is the same as in training and will remain so, that boundaries between episodes are absolute, that agents are by definition indifferent to impacts of their actions over long time-horizons, and so on. In short, RL (and ML more broadly) most often seeks optimal solutions to fixed problems. We later address (in Section \ref{sec:openworld}) extensions of RL (such as meta-learning) that in theory, but (we argue) not in practice, meet this critique.
Secondly, RL research is often driven by implicit assumptions about how existing agendas will naturally lead to superhuman robustness. For example, that the combination of increasing scale and the generalization capabilities of large neural networks will address all relevant robustness issues; or that it will do so when combined with increasingly sophisticated hand-designed RL algorithms derived to fit closed-world formalisms. Yet it may be optimistic that robustness to unknown unknowns will arise as byproducts of formalisms and methods that implicitly disavow their existence.

\begin{figure}[h!]
    \centering
    \includegraphics[width=0.65\textwidth]{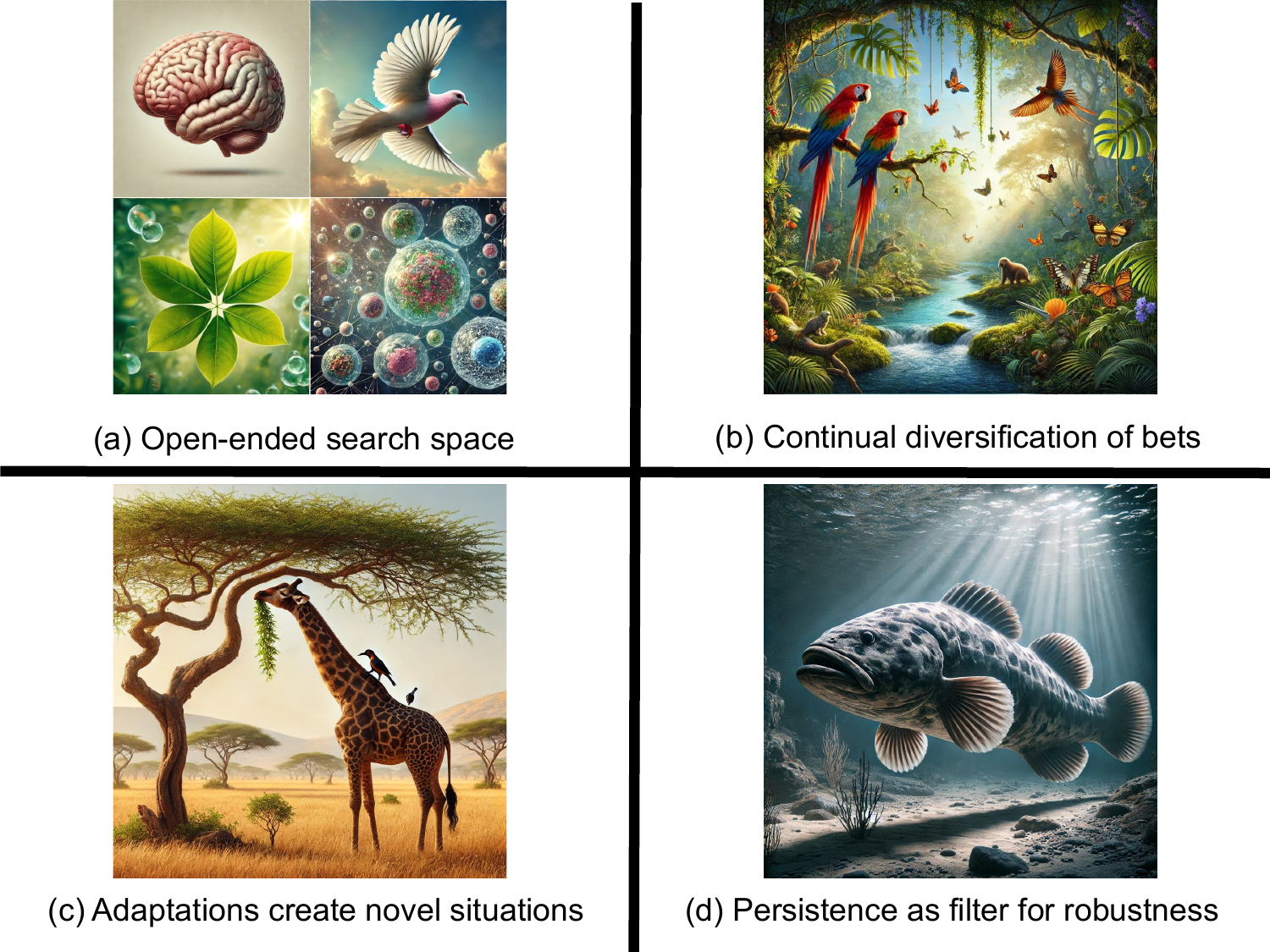}

    \caption{\textbf{Interlocking principles enabling evolution's robustness to Knightian Uncertainty.} (a) Evolution happens within a search space that is open-ended enough such that a vast array of complex adaptations can be encoded, e.g.\ the human brain, multicellularity, developmental systems as a whole, and photosynthesis. (b) Diversification pressure in evolution continually creates new behaviors and adaptations from the set of open-ended possibilities, which implicitly can be seen as bets about how the organism and its lineage can persist into the future. (c) Because organisms form part of the environment of other organisms, novel behaviors and adaptations in one lineage create novel unforeseen situations for other organisms as an externality, e.g.\ the high branches of a tree provide a novel situation a giraffe can exploit. (d) Organisms unable to persist across the uncertainty created by other organisms are filtered away, in effect invalidating their bets about how to persist through KU; the image shows a coelacanth, a fish that has persisted for 400 million years. In concert, these factors can be seen as a form of open-ended generation and falsification of bets about how to deal with KU. We believe there may be ways to adapt these principles to ML research (see discussion in Section \ref{sec:discussion}).}
    \label{fig:diagram}
\end{figure}

\begin{figure}[th!]
    \centering
    \includegraphics[width=0.7\textwidth]{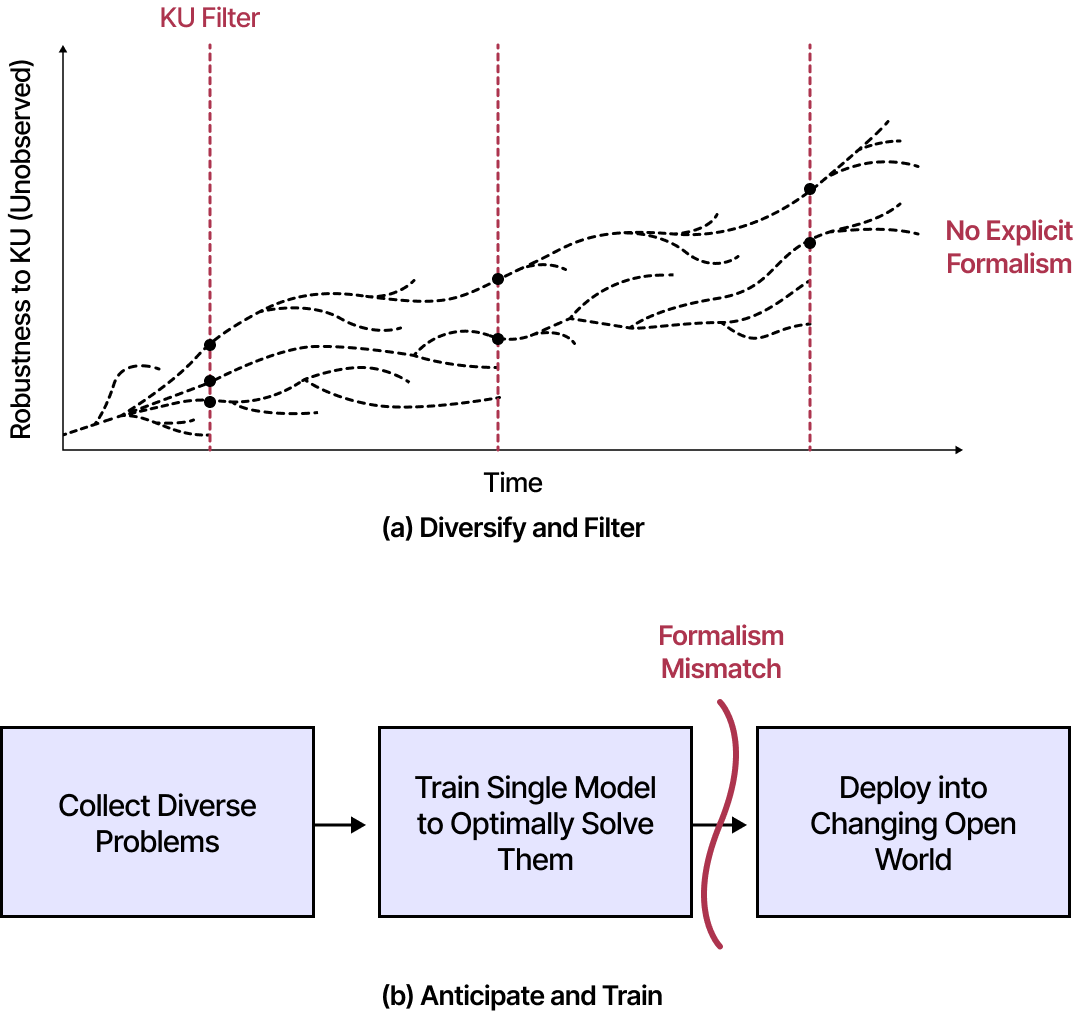}

    \caption{\textbf{Two Strategies for Dealing with an Open World.} This figure describes two possible strategies for coping with an open changing world. In (a) \emph{diversify-and-filter}, a process continually refreshes and adapts its diverse hypotheses about how to persist through the open-ended future. Such hypotheses are filtered through empirical success at tackling later unanticipated problems. Evolution, market competition, and science can be seen to largely operate through this paradigm. There is no explicit formalism, although robustness implicitly relies on the Lindy effect \cite{taleb2014antifragile}, i.e.\ an adaptable solution long-tested by time is more likely than an untested one to persist yet longer. 
    In (b) \emph{anticipate-and-train}, diverse problems are first collected, and augmented through human anticipation about what novel situations might later arise. Then, a single policy is trained to solve the problems to convergence, and that policy is then deployed into a changing world. Much of current ML adopts this paradigm; although the closed-world formalism adopted in training mismatches the open world it is deployed into, the hope is that generalization will enable sufficient robustness to unforeseen challenges. 
    One conclusion is that nothing precludes machine learning from more deeply integrating diversify-and-filter approaches into its methods \cite{lehman2010revising,kumar2020one,jaderberg2017population,lee2023diversify}. Another conclusion is that diversify-and-filter 
    leverages the temporal structure of when novel problems arise, and forces agents to directly grapple with the issue of KU (if they do not, they are discarded).}
    \label{fig:diversify}
\end{figure}

In contrast, biological evolution allows all aspects of an agent's architecture to adapt; nothing is sacrosanct. There are no explicit formalisms underpinning the many divergent implementations of biological reinforcement learning, nor any fixed time horizon beyond which feedback cannot shape evolution's trajectory. Evolution continually seeks new and diverse ways of surviving and reproducing, which implicitly represent bets on the nature of the future (e.g.\ perhaps dinosaur-sized reptiles are robust; or small mammals); this diversification can continually refresh the many bets culled by reality (with no fixed time horizon), such that time favors bets (lineages) more robust to the continually unfolding unknown (which itself is partially created by the diversification of life, in how one species contributes to the environment of others). Figure \ref{fig:diagram} highlights these conditions through which evolution becomes robust to KU; and figure \ref{fig:diversify} contrasts at a high level how evolution and ML approach robustness to the open unknown.

One possibility is that, similar to the argument for AI-generating algorithms (AI-GAs; \cite{clune2019ai}), there remains \emph{further} bitter lessons \cite{sutton2019bitter} that ML has yet to absorb. Dealing with KU is central to intelligence and remains a fundamental challenge for ML, and to achieve it, computation may trump clever combinations of individual algorithmic advances. That is, it may be more effective to scale simpler algorithms that make fewer assumptions, taking inspiration from the genesis of robust human learning, i.e.\ biological evolution. That ``more [computation] is different'' \cite{anderson1972more} extends beyond deep learning alone \cite{such2017deep,salimans2017evolution}, and less attention has been placed upon fields such as artificial life \cite{adami1998introduction,langton1997artificial} that while highly ambitious, have revolutionary potential.

Or conversely, perhaps there are ways to tackle KU through novel RL algorithms that deeply integrate insights from evolution or how humans and society successfully navigate KU \cite{kay2020radical,taleb2014antifragile,johnson2023conviction}. In total, this paper adds to a thread of counter-intuitive critique, that despite dramatic recent progress, ML and RL may yet still be skirting a core feature of intelligence \cite{chollet2019measure,samothrakis2024games,mitchell2021ai,hughes2024open}. 
Importantly, while this paper presents a critique of ML, we are not skeptics of algorithmic intelligence nor dogmatically committed to mimicking biological evolution or the primacy of evolutionary algorithms; instead, we aim to draw focus to a blind spot of ML, in service of moving its formalisms and algorithms forward.

The paper proceeds first by reviewing background on KU and on the tension between ML and open worlds. Next, Section \ref{sec:bioevo} explores the means through which biological evolution is able to encourage robustness to KU is explored, and Section \ref{sec:rllimits} contrasts them with the formalisms of RL. Finally, Section \ref{sec:discussion} discusses the possibilities for the fields of artificial life and open-endedness to contribute to reckoning with unknown unknowns; implications of KU for large models and AI safety; and on possibilities for more deeply integrating KU into RL.

\section{Background}

The next sections review (1) the idea of Knightian uncertainty, a central concept in this paper, and (2) the existing ways in which the field of ML wrangles with the challenges of an open world.

\subsection{Knightian Uncertainty}
\begin{quote}
`` Before the wheel was invented [...] no one could talk about the probability of the invention of the wheel, and afterwards there was no uncertainty to discuss [...]. To identify a probability of inventing the wheel is to invent the wheel.''
\hfill --- John Kay and Mervyn King \cite{kay2020radical}
\end{quote}

\begin{quote}
``What gets us into trouble is not what we don't know. It's what we know for sure that just ain't so.'' 

\hfill --- Mark Twain
\end{quote}

\begin{figure}[h!]
    \centering
    \includegraphics[width=0.85\textwidth]{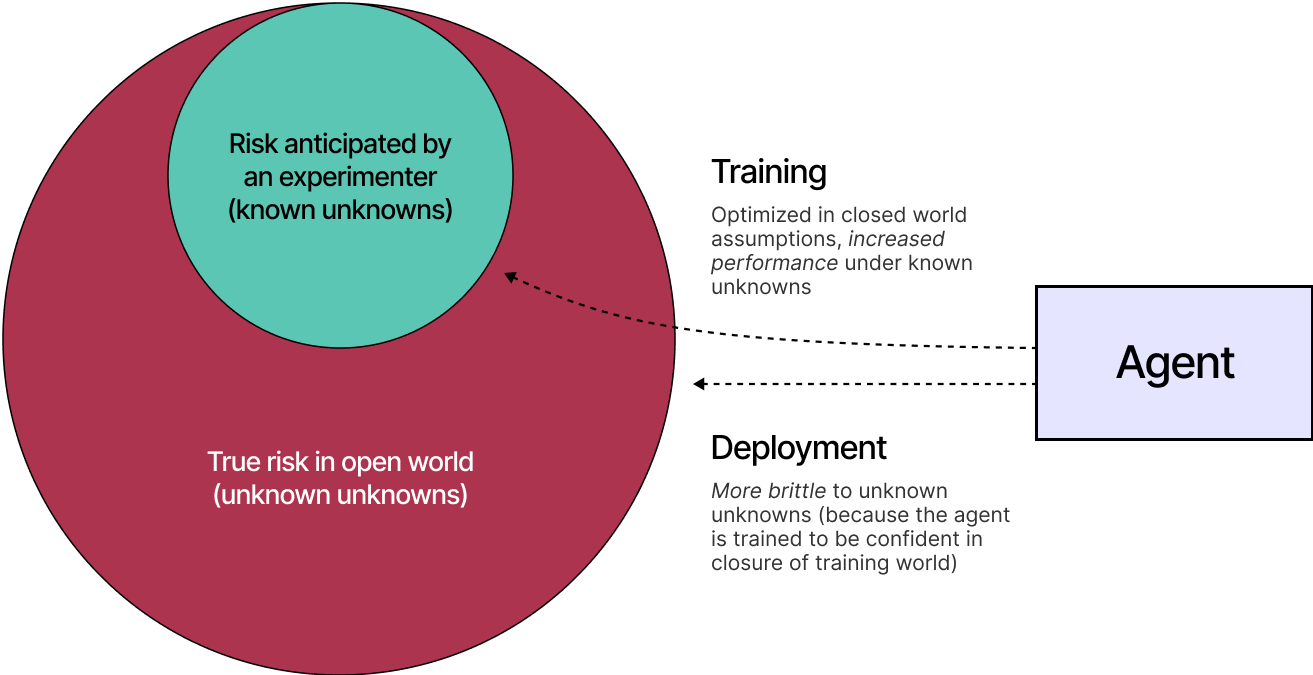}

    \caption{{\textbf{Optimizing for known unknowns can exacerbate risk from Knightian uncertainty.}} An optimization formalism that makes closed-world assumptions will indeed improve an agent's performance on the situations an experimenter anticipates. However, if such a \emph{closed-world} optimizer aggressively trains an \emph{open-world} agent, the agent may perversely become \emph{more brittle} to Knightian uncertainty, as it is incentivized to internalize the closed-world assumptions as true.}
    \label{fig:truerisk}
\end{figure}

The attempt to manage risk through formalizing it is common practice in many fields such as finance \cite{christoffersen2011elements}, economics \cite{mongin1998expected}, and machine learning \cite{moos2022robust,suresh2008risk}. Yet when formalizing risk and optimizing against it, we can delude ourselves into a false sense of security by thinking we have the world pinned down \cite{kay2020radical}. Much escapes formalization, however, and in optimizing against what we formalize, we may indeed decrease the narrow form of quantified risk, while potentially exacerbating true risk. Figure \ref{fig:truerisk} provides a high-level depiction of this phenomenon.

In other words, Goodhart's law applies: Optimization pressure decouples a simplified metric from the broader quality it is intended to capture \cite{goodhart1984problems,stanley2015greatness}. 
A prominent example is the subprime mortgage crisis of 2007, which at least partially resulted from incorrectly assuming away long-tail risks (in particular, assuming that defaults on high-risk loans could not become highly-correlated), and optimizing against that faulty model with abandon, as if true risk were contained \cite{mackenzie2014device,kay2020radical}. 

It is likely impossible to anticipate all important aspects of the future given the complexity of the world, which grows daily through the decentralized actions of billions of creative agents, who create new products, organizations, media, software, technology, ideas, etc. We can model known unknowns, i.e.\ the things we know we do not know -- but by definition cannot predict unknown unknowns, i.e.\ what we do not know that we do not know (see Table \ref{table:risk}). Formalizing the idea of unknown unknowns is near-paradoxical, and the closest concept is that of Knightian uncertainty in economics, which was independently posed in 1921 by John Maynard Keyes and Frank Knight \cite{knight1921risk,dimand2021keynes}. Knight makes a distinction between what he calls uncertainty and risk: ``Uncertainty must be taken in a sense radically distinct from the familiar notion of Risk ... a \emph{measurable} uncertainty, or `risk' proper, as we shall use the term, is so far different from an \emph{unmeasurable} one that it is not in effect an uncertainty at all \cite{knight1921risk}. [emphasis added]''

\begin{table}[h!]
\centering
\renewcommand{\arraystretch}{1.5}
\begin{tabular}{|c|c|c|}
\hline
\textbf{} & \textbf{Knowns} & \textbf{Unknowns} \\ \hline
\textbf{Known}   & Deterministic (Known Knowns) & Risk (Known Unknowns) \\ \hline
\textbf{Unknown} & Implicit Knowledge (Unknown Knowns) & Knightian Uncertainty (Unknown Unknowns) \\ \hline
\end{tabular}
\vspace{0.1in}
\caption{{\textbf{Types of Risk and Uncertainty}.} Optimization is most straightforward in deterministic settings with complete information, and more challenging in situations of risk, where there are unknowns, but their parameters are known or estimable. More challenging still are situations where there exist unknown unknowns, i.e.\ future possibilities that are qualitatively novel and difficult or impossible to anticipate (though they may seem obvious in hindsight) -- the central topic of this paper. The fourth quadrant of unknown knowns (e.g.\ where an agent or system has implicit knowledge that is not explicitly known) is also important but less relevant to arguments here. In short, machine learning excels at situations of determinism or risk, but struggles to model Knightian uncertainty.}
\label{table:risk}
\end{table}

Note that Knight's usage of ``uncertainty'' differs from common usage in ML. Rather than uncertainty in Bayesian statistics \cite{bishop2006pattern}, Knightian uncertainty intuitively relates to unknown unknowns, where some important outcomes cannot be probabilistically quantified or anticipated. Such KU contrasts with what Knight calls ``risk,'' wherein uncertainty can be modeled (i.e.\ known unknowns). ML's treatment of uncertainty rarely contends with unknown unknowns; Bayesian methods require, for instance, defining the space of possible hypotheses to calculate uncertainty across \cite{ghavamzadeh2015bayesian,johnson2023conviction} and are subject to restrictive closed-world formalizations in the same way as ML more broadly 
(discussed later in Section \ref{sec:rllimits}).
A recent work has also highlighted KU as an important property for AI \cite{samothrakis2024games} in the context of games; and other recent critiques of ML and RL can be interpreted from the lens of KU \cite{chollet2019measure,abel2024three}.

Because KU makes a negative claim (i.e.\ it deals with situations where uncertainty \emph{cannot} be modeled), it is understandably a controversial concept \cite{coddington1982deficient,dimand2021keynes}. While the idea of unknown unknowns is intuitive, as each of us has been buffeted by events we reasonably failed to anticipate, it cuts against many fields' tendencies to create formalisms that by fiat rule them out as possibilities, e.g.\ neoclassical economics, game theory, and subfields of machine learning. For example, the formalization of supervised learning assumes a test distribution identical to that in training \cite{cao2016non,gu2021beyond}. In such a frozen world, there is no room for unknown unknowns;
yet assuming away their existence need not make them less real or important.

As a representative example of a KU skeptic, the economist Milton Friedman wrote ``I have not referred to this distinction [between Knight's risk and uncertainty] because I do not believe it is valid. [...] We may treat people \emph{as if} they assigned numerical probabilities to every conceivable event \cite{friedman1976price}. [emphasis added]'' In other words, as in many economic formalizations \cite{becker1976economic,savage1972foundations,lucas1972expectations}, Friedman \emph{assumes} that all uncertainty can be framed probabilistically, and that effectively the agent should be blamed for not anticipating all emergent possibilities in an open world (regardless of computational intractability). A similar assumption is often made within ML, but with blame instead implicitly assigned to the experimenter who fails to anticipate all future scenarios when designing the training environment. Limits of anticipation of course exist: Taken to an extreme, what basis does a caveman have for estimating the year in which GPUs will be invented, lacking even the concept of a computer at all? We do often find ourselves in situations of true uncertainty (what are the long-term consequences of taking this job rather than the other? of having a child? of starting this business? of moving to a new city?), and the qualitative distinction between business as usual and situations of true uncertainty matters (see longer counter-arguments in \cite{sunstein2023knightian,kay2020radical}).

\begin{figure}[h!]
    \centering
    \includegraphics[width=0.8\textwidth]{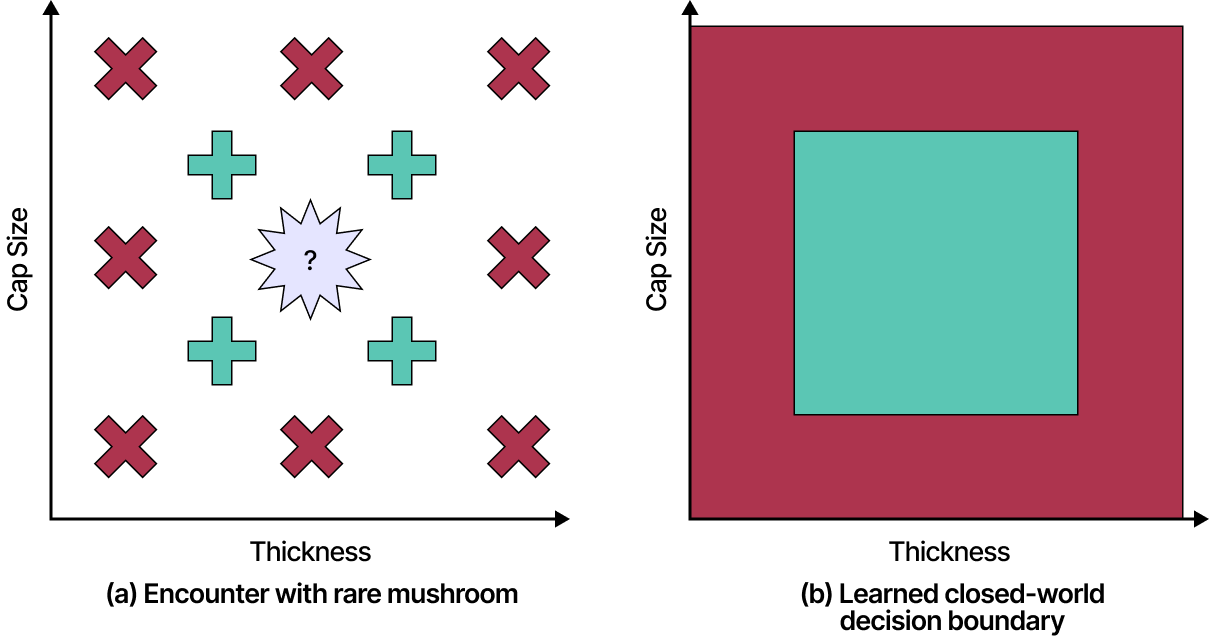}
    \caption{{\textbf{Neural network generalization is not a general cure for Knightian Uncertainty.}} Imagine as part of a larger reinforcement learning policy, an agent decides whether to eat certain mushrooms, which can either be deadly or edible, and can be separated through features learned in training that correspond to the cap size of the mushroom and its thickness. (a) In a closed world, it is safe to assume that the distribution of mushrooms encountered during training (red $\times$'s and green $+$'s) reflects that encountered during testing, and the (b) NN decision boundary on whether to eat or not eat the mushroom learned through training will likely reflect this assumption. However, in an open world, not all mushroom varieties are known, the policy might be deployed in a slightly different ecosystem, or a new variety of mushroom might evolve or be bred. If encountering the unanticipated mushroom (question mark symbol) at the center of (a), it is likely rational for an open-world agent to forgo eating it, given its novelty and the risk of death. The claim is that simple generalization from what is known does not address Knightian uncertainty.}
    \label{fig:mushroom}
\end{figure}

In ML terms, a neural network (NN) policy can of course, given any possible input, output a probability distribution across actions. Such a policy is trained over some experimenter-provided distribution of training situations. When deployed, and facing an unknown unknown, one can query the network to output a probability. Yet, there is no theoretical reason to assume sensible generalization from the underlying function approximator (e.g.\ NN) when querying it far out-of-distribution. More broadly, it is wishful thinking to assume
that the best policy for a qualitatively novel situation is an interpolative generalization from previously-encountered situations. For example, it is better not to eat a mushroom species you have never encountered before, even if it exhibits combinations of features shared by mushrooms you know to be edible (see Figure \ref{fig:mushroom}). Subtle differences in input can often be of outsize importance when generalizing, meaning that uncertainty quantification methods may not mitigate this failure \cite{abdar2021review}, especially when the situation violates the assumptions of the training algorithm (to be discussed more in the next section). Note than even if uncertainty quantification methods reveal that the situation is novel, \emph{what to do} to gracefully handle such a situation can be both high stakes and highly context-dependent (i.e.\ not a simple matter of generalization).

This paper thus takes as given that the distinction between risk and KU is meaningful, especially as it highlights  distinctions between algorithms that treat risk explicitly in terms of formal probabilities, e.g.\ the expected value under a stochastic policy, and those that do not (e.g.\ evolution as a whole, or an animal learning to deal with a new human invention). One might then concede that KU exists, but remain skeptical of its practical implications, i.e.\ whether \emph{any} algorithm could handle it with grace; perhaps in situations of true uncertainty  there are no reasonable principles of action. 
Yet in practice some decision policies are indeed more fragile than others in unexpected situations \cite{kay2020radical,taleb2014antifragile,johnson2023conviction,knight1921risk}. For example, while in normal conditions it may be of marginal advantage for an airplane to carry only enough fuel to reach its destination, under many unexpected situations, having additional fuel will allow more flexibility to handle them without catastrophe. Similarly, Nassim Taleb highlights the benefits of antifragile systems \cite{taleb2014antifragile}, i.e.\ systems that become stronger in the aftermath of unpredictable but impactful events (black swan events; \cite{taleb2010black}).
Indeed, Taleb describes biological evolution as a prototypical example of a system that benefits from unexpected shocks. We will return to the theme of evolution later, after next reviewing the challenge of ML in open worlds.

\subsection{Machine Learning in Open Worlds}
\label{sec:openworld}

\begin{quote}
    ``There are more things in heaven and earth, Horatio, / Than are dreamt of in your philosophy.''

\hfill --- William Shakespeare
\end{quote}

The field of ML most often assumes a \emph{closed world}, where representative situations are provided in a training distribution, and that the world will not thereafter change \cite{zhu2024open}. That is, the dominant abstractions in supervised and unsupervised learning assume that data is independently and identically distributed (IID; \cite{cao2016non,gu2021beyond}). A similar assumption in most RL formalisms is that the deployment environment is the same as the training environment \cite{zhang2018study,kirk2023survey}.

However, the world in fact is open-ended, ever-changing, and presents many rare situations that may effectively be impossible to fully anticipate a priori \cite{popper2005logic,hayek2013use,macmillan2023ir,dequech2001bounded,longo2012no}. An open world entails unknown unknowns, a fact sometimes embraced in other fields of study, such as economics, complexity theory, business strategy, and risk management \cite{taleb2010black,knight1921risk,arthur2021foundations,rindova2020shape,longo2012no}, but rarely directly confronted within ML \cite{chollet2019measure,hernandez2017evaluation}. The divergence between strong formal assumptions in ML and real-world situations is well-known and matters in practice \cite{farahani2021brief,zhou2022domain,vanschoren2018meta,novak2018sensitivity}, as it is a central cause for the challenge of real-world ML deployments. As a result, there are many strategies for relaxing the assumption of a closed world or directly improving the robustness of ML solutions, which are surveyed next.

\subsubsection{Scaling and Generalization}

 For example, accumulating training data can increase coverage of unlikely situations. Such accumulation, coupled with NN generalization, does indeed lead to surprising and powerful capabilities \cite{brown2020language,wei2022emergent}. Yet as impressive as such large models are, they still are susceptible to jail-breaks (novel adversarial inputs), hallucination of references (an interesting failure of robustness), and surprising generalization failures \cite{chollet2019measure,wu2023reasoning}. In short, relying on the capabilities of NNs to generalize beyond their training distribution, no matter how much data is accumulated, does not yet seem to solve the problem of robustly dealing with unknown unknowns (see also Figure \ref{fig:mushroom}, and later discussion in Section \ref{sec:generalization}). We fully acknowledge that the failures of current models are a moving target, and that
 scaling has proven an incredibly powerful paradigm; at the same time, the arguments in this paper highlight a subtle
 but important facet of robust behavior that scaling alone seemingly does not address.

\subsubsection{Meta-learning and Continual Learning}

\begin{figure}[th!]
    \centering
    \includegraphics[width=0.85\textwidth]{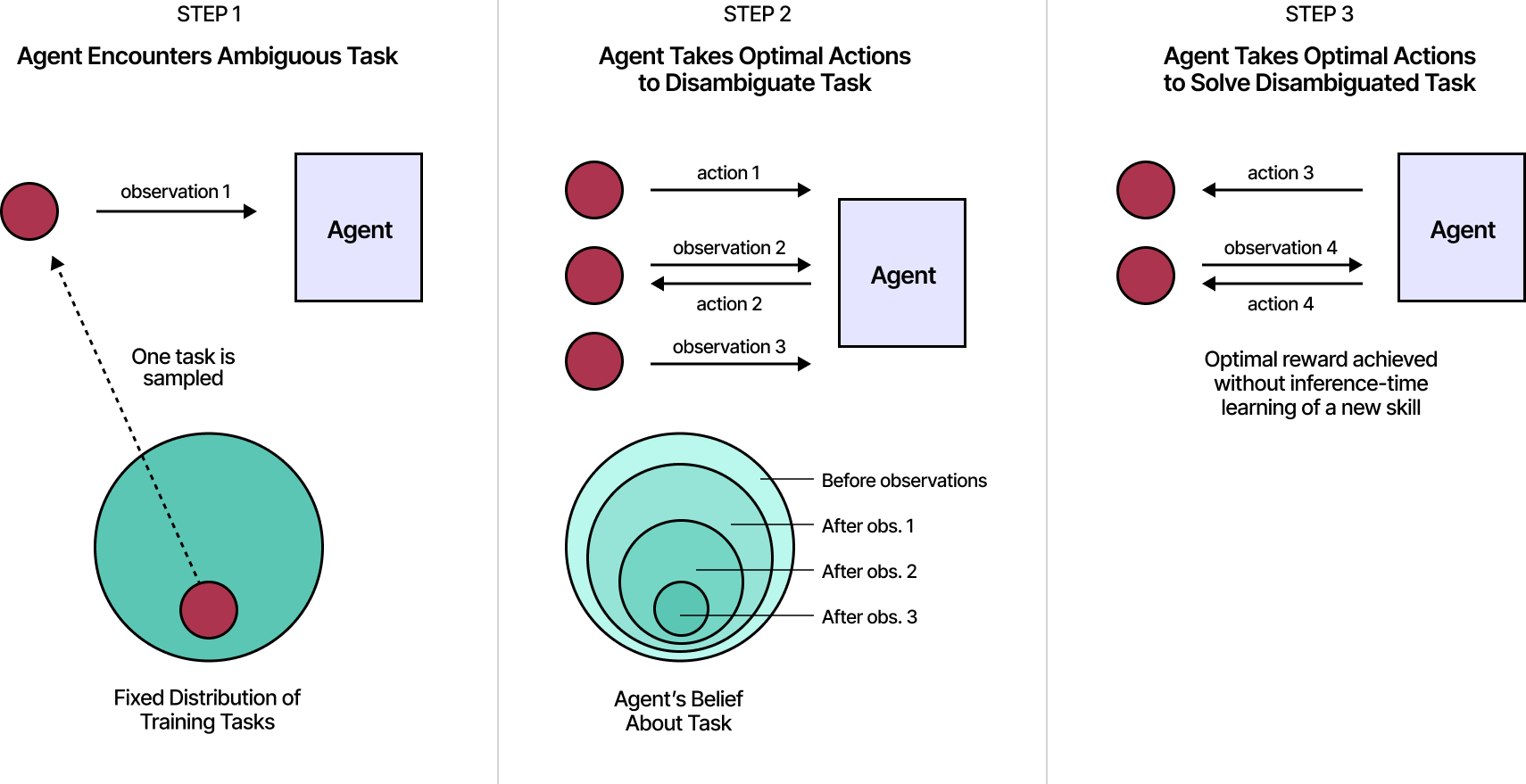}
    
    \caption{\textbf{Typical metalearning setups do not incentivize learning how to solve unforeseen tasks.} This figure offers a caricature of optimal behavior under a typical meta-RL formalism, where an agent is trained across a fixed distribution of problems; this setup is similar to e.g.\ \cite{duan2016rl}. In meta-RL, it is common for an agent to be exposed many times to training tasks covering all major necessary task-relevant skills. Thus the agent is incentivized to learn \emph{in training} all qualitative skills needed to solve the tasks; after many iterations of training, there need be no significant remaining surprise for the agent when solving new tasks drawn from an IID test distribution (which is the formal goal of the algorithm). At completion of training, an optimal agent's behavior is sketched as: (1) it encounters a task drawn from the IID test distribution, which is ambiguous (as this characterizes the need for metalearning); (2) the agent takes actions that optimally disambiguate the sampled task; and (3) having identified the task, which it has encountered many similar variants of before in training, the agent executes its previously-learned optimal solution. In practice, optimal behavior will entail mixing steps (2) and (3) together, but nowhere in this process does optimality under the formalism require the generalized ability to learn how to learn. The conclusion is that if then deployed into a changing world where it encounters an unknown unknown, the agent may struggle to handle it gracefully.}
    \label{fig:metalearn}
\end{figure}

Other mitigation strategies include explicitly relaxing the assumption of a static data distribution, through meta-learning \cite{vanschoren2018meta} or continual learning \cite{parisi2019continual}. Meta-learning, especially when called ``learning how to learn,'' appears a natural paradigm for handling unknown unknowns; however, it is worth examining what actually is entailed by common meta-learning formalisms. In practice, the assumption of IID, characteristic of closed worlds, is shifted from a static training distribution to a meta-distribution of possible tasks. The world remains closed, but the researcher is now responsible for specifying that meta-distribution, and there is no reason that the learned policy should generalize to qualitative variants of tasks the researcher fails to anticipate \cite{rajendran2020meta,kirsch2019improving,ajay2022distributionally}. As a representative example, Figure \ref{fig:metalearn} highlights how optimal behavior in meta-RL can meaningfully differ from commonsense notions of ``learning how to learn.''
Continual learning makes distinct but related assumptions \cite{wang2024comprehensive,parisi2019continual,khetarpal2022towards}, focusing more centrally on the challenge of catastrophic forgetting rather than of robustness to novel situations; although some frame it as inclusive of meta-learning \cite{khetarpal2022towards}. In conclusion, both meta-learning and continual learning are exciting directions, although neither approach (to our knowledge) has successfully tackled many qualitatively new situations across long deployments; nor do such methods play much role in current foundation models deployments (with exception of in-context learning, which can be seen as a form of emergent meta-learning \cite{dai2022can}).

\subsubsection{Safe Reinforcement Learning}

Another family of mitigations focuses more directly on the issues of risk and robustness. For instance, the area of robust reinforcement learning \cite{moos2022robust} aims to discover policies for agents that can better cope with uncertainty, shocks, and change. Nearly all such methods rely on \emph{formalizing} robustness through making \emph{quantitative} assumptions about how a reward function or state transition dynamics might change, or what kind of observation or action noise an agent might encounter; while effective and useful for closed world situations, the challenge in open worlds relates more to qualitative changes. Indeed, recent work has suggested that common framings of RL under risk are insufficient to deal with black swan events \cite{lee2024black}. Most such approaches implicitly assume that \emph{qualitative} changes in e.g.\ state transition dynamics (like a new type of actor in the world) can be handled through adding \emph{quantitative} noise to the system; and it is left to the experimenter to decide what kind of quantitative robustness to encourage.

\subsubsection{Open World Recognition}
A related direction focuses on open set \cite{bendale2016towards} and open world recognition \cite{bendale2015towards}, wherein it is acknowledged that a classification system will encounter unknown unknowns corresponding to unanticipated categories beyond those upon which it was trained; such systems often leverage meta-recognition \cite{zhang2014predicting} algorithms that attempt to predict when a system is likely to make an error. While such algorithms aspire to \emph{detect} novel situations, and more directly confront the issue of unknown unknowns, they do not address the crucial issue of \emph{how to robustly respond to them}. Such algorithms are indeed interesting and worthy of more interest, and may provide a promising starting point for greater integration of KU-awareness into ML.

\subsubsection{Zero-shot Generalization in RL}

A setting very relevant to the one described in this paper is zero-shot generalization (ZSG) in RL \cite{kirk2023survey,higgins2017darla}; although not framed in terms of unknown unknowns, the ability to generalize immediately to environments unlike those handled in training is similar to KU robustness. However, to our knowledge current ZSG methods do not yield robustness to qualitatively new challenges or degrees of variation. Additionally, the formalism of ZSG does not take into the account the temporal process through which new situations emerge in an open world, which we argue is key to how real-world agents handle KU (see Figure \ref{fig:diversify}). Yet this is an exciting area of research, and may help meet the challenge of KU.

\subsubsection{Open-endedness}

Finally, the field of open-endedness \cite{soros2017open,stanley2015greatness} attempts to study and engineer processes of open-ended creativity, often inspired by biological evolution. The idea is to create long-running search processes that generate continual innovation, as happens in culture and science. Such open-endedness aims in effect to create and continually \emph{contribute} to an open world, and work has explored how evolutionary open-endedness leads to increasing evolvability \cite{lehman2011improving,lehman2016critical}, i.e.\ the ability for the system or individuals to quickly adapt. However to our knowledge, no work has attempted to directly tackle the problem of KU through open-endedness, although it has been applied to increase the capacity and adaptability of RL policies in impressive ways 
\cite{dennis2020emergent,team2021open,team2023human,wang2019paired}, most often within a fixed class of environment where tasks require relatively short time horizons.

\subsubsection{Summary} In general, across most such approaches, there is an intriguing tension between the desire to \emph{derive} algorithms from formal problem definitions and the seeming \emph{impossibility} of formalizing the ways in which the future will \emph{qualitatively} differ from the present (and how to respond to such qualitative differences). The proliferation of problem settings, whether meta-learning, continual learning, stochastic games, robust RL, gives the impression that ML is marching towards solving its deepest issues, if only enough progress is made in each subproblem and then combined \cite{clune2019ai}, yet it is also possible that little research actively targets what is core: The past does not in general provide a straight-forward guide to the future \cite{stanley2015greatness}.

\section{Biological Evolution and Knightian Uncertainty}
\label{sec:bioevo}

\begin{quote}
``Darwin’s idea is a universal acid; it eats through just about every traditional concept, and leaves in its wake a revolutionized worldview.''

\hfill --- Daniel Dennett
\end{quote}

This paper argues that the formalisms of ML have a blind-spot for unknown unknowns. This section proceeds through proof-by-example that greater robustness is possible through different mechanisms: Biological evolution, though it unfolds without volition or formalisms, does indeed create organisms (such as ourselves) that are surprisingly robust to novel and outlier situations. This section argues that evolution's products are often robust to KU, and highlights the mechanisms through which such robustness arises. The subsequent section then examines limiting assumptions made in RL, the ML paradigm most relevant for learning to handle the unknown.

To begin, we sometimes  take for granted the incredible complexity and surprising robustness of biological evolution's products, especially given the dearth of information that drives its search. To a first approximation, evolution blindly pulls mutational levers and receives as feedback only whether the levers pulled result in persistence across time \cite{grand2001creation}. Evolution has no foresight, makes no formal assumptions, and has \emph{no conceptual understanding} of the complex mechanisms through which an organism survives and reproduces. And yet its creations often act \emph{as if} they are making complex plans about the future: e.g.\ some cicadas have life cycles that synchronize to prime numbers (e.g.\ 13 or 17 years), thereby making it more difficult to be predated by organisms with shorter lifecycles (as prime numbers are not divisible by smaller numbers); leafcutter ants in effect farm fungi by cutting leaves (too tough for them to eat) that they feed to underground symbiotic fungal gardens; migratory birds can travel thousands of miles to exploit seasonal food resources; some pine species have cones that lie dormant and only release seeds after being exposed to fire, ensuring that the environment is cleared of competition and nutrient-rich; and cyanobacteria living only hours can keep track of changing seasons across generations \cite{jabbur2024bacteria}. 

To quote Cass Sunstein \cite{sunstein2023knightian}: ``Contrary to a standard view in economics, Knightian uncertainty is real. Dogs face Knightian uncertainty; horses and elephants face it; human beings face it [...]'' to which we could add: biological evolution most certainly faces it. Evolution is fundamentally unable to assign explicit probabilities to novel outcomes, and yet seems also to be remarkably successful at navigating uncertainty. On the whole its creations are surprisingly robust in the face of unforeseen challenges \cite{kitano2004biological,roberts1980robustness}; and more deeply, evolution as a whole is incredibly robust -- it is unlikely to be extinguished on Earth, even given full-scale nuclear war.

As a consequence of its lack of volition, evolution necessarily must wrangle with KU, in contrast with ML algorithms derived in explicitly probabilistic terms from closed-world formalisms. There is no theoretical formalism of risk that evolution imposes upon the world; instead, it can be viewed an ongoing anarchic, creative, and divergent search for interdependent structures that persist across time, subject to the stringent constraints of the environment and interactions with other organisms.

\definecolor{brown}{RGB}{0, 120, 0} 
\definecolor{red}{RGB}{180, 0, 0} 

More concretely, the argument is that evolution's robustness in the face of KU is due to four interlocking factors (see also Figure \ref{fig:diagram}; note that these principles relate to previously-proposed conditions for enabling continuing open-ended evolution \cite{soros2014identifying}, but aim instead to explain the emergence of robustness to KU):

\begin{enumerate}
\item {\color{brown}The selection criterion of persisting across (potentially vast) swaths of time filters bets for robustness.}
\item {\color{violet}The drive to accumulate an abundant diversity of solutions results in many diverse bets on how to persist through future.}
\item {\color{red}Adaptations of one organism create novel situations for others that tests their robustness in new ways.}
\item {\color{blue}An open-ended search space enables revising any part of the mechanisms of learning in service of greater robustness.}
\end{enumerate}

In short, an organism implicitly encodes a bet about how it and its offspring can {\color{brown}persist through the indefinite future}, which is tested across unforeseeable situations generated in part by the physical environment and in part by the {\color{red}adaptations of other organisms}. Because {\color{violet}evolution explores a diverse and ever-refreshing set of such bets}, and {\color{brown}bad (or unlucky) bets are culled across both the short and long-term}, as a whole it tends to favor organisms relatively robust to the unforeseen. Further, because {\color{blue}the search space includes arbitrarily complex adaptations}, including revising any aspect of neural architecture, learning algorithms, and development as a whole, {\color{blue}there is no commitment in the system to any particular formalism, algorithm, or architecture}. Instead, {\color{brown}reality serves as a filter for what works empirically} in the face of often-extreme uncertainty. 

To expand on this argument, we next connect evolution to Popperian falsifiability in science (to highlight how evolution filters losing bets), and also expand on evolution's open-ended drive towards diversity (to highlight how its generates both new bets and novel situations for testing them). Readers with less interest in evolution can proceed directly to Section \ref{sec:rllimits}; what is most critical is that the mechanisms described above could plausibly be adapted and abstracted into ML algorithms as means of encouraging KU robustness.

\subsection{Evolution as Driven by Scientific Falsifiability}
\begin{quote}
%
``We are standing in the doorway of a library of life, but the books are burning as we speak.''

\hfill --- GPT-4o hallucination attributed to E.O. Wilson
\end{quote}


Popperian falsifiability is the idea that a scientific theory must be testable in a way that could prove it false \cite{popper2005logic}, and provides a useful frame for thinking about what it means for an evolutionary lineage to {
persist across (potentially vast) time}. The scientific community creates many divergent hypotheses across its diverse fields of study (e.g.\ physics, psychology, biology), which can be falsified through experiments that generate results that cleave between hypotheses (e.g.\ Francesco Redi's disproval of the spontaneous generation theory within biology \cite{redi1909experiments}). We can similarly view the diverse gambles of evolution from the lens of scientific hypotheses, experimentation, and falsification \cite{popper2005logic,dennett1995darwin}. An organism’s current genome represents a range of possible phenotypes (expressible through development and environmental influence); this range of possibilities can be viewed as a hypothesis about how an organism can persist the future, which can be falsified through ``experiments'' with reality if it is not able to survive and reproduce. This analogy with science is helpful, because it clarifies that evolution is in effect {
an automated and diversifying research and development process} \cite{soros2017open}, and that like science, it pursues many independent tentative bets which at all times are subject to falsification, {
no matter how long they have so far persisted}.

Importantly, beyond a scientific paradigm's ability to explain current experimental data, its longevity depends also on its potential to be fleshed out and expanded in the future -- the extent to which it is a generative stepping stone \cite{stanley2015greatness}. For example, the beauty of Darwin's insights into evolution persist today, through being much-expanded and refined into the current extended evolutionary synthesis \cite{pigliucci2010evolution}; 
in a similar way, a genome is a stepping stone to a wider range of possible future phenotypes accessible through genetic mutation, i.e.\ its adjacent possible \cite{kauffman1993origins}. 
In other words, persistence of an organism's lineage depends not only on the behaviors it exhibits during its life (and their robustness to KU), but to its potential for future adaptation and diversification (e.g.\ to realize adaptive radiation \cite{stroud2016ecological}), as new situations and opportunities emerge across generations.


\subsubsection{Bets on Robust Behavior in an Organism's Lifetime}

Biological organisms have evolved diverse strategies for robustly navigating uncertainty during their lifetimes. For example, many animals have developed avoidance strategies that are robust through their simplicity, e.g.\ toads that attempt to eat small moving things, and avoid large moving things \cite{ridley1995animal}; and mice have a generalized fear of open spaces where they are vulnerable \cite{lister1987use}. Mammals more generally have a variety of sophisticated fear and fear-learning mechanisms \cite{ohman2001fears}, often biased towards rapid response (e.g.\ that relies on simple visual features), and favoring false positives over false negatives, given the life-and-death stakes of predation. Notice in this example, that {
evolution moves outside the abstractions generally available to a typical NN approach to RL}: It leverages a faster response to a dangerous situation by dynamically using less neural processing. Such rapidity is a fit for avoiding predators, and less rapid responses are culled by evolution as they result in greater predation. The point is not rapid response per se, but that in general, {
evolution benefits from lacking a firm commitment to any particular ingredient or formalism} (unlike a single NN that follows a fixed path of computation from input to output).

\subsubsection{Bets on Robust Behavior in an Organism's Future Lineage}

A lineage's evolutionary success extends far beyond the lifetime of one organism, {
requiring continuing persistence over long swaths of time} \cite{grand2001creation}. Any organism alive today has an unbroken track record of persistence dating back to the first replicator, across which its ancestors undoubtedly successfully met the challenges of countless unanticipated situations. If an organism's genome is overly-optimized for expected present reproductive success, but its entire lineage falls prey to a catastrophic change in conditions, then that bet on the future is culled. Thus, even if an organism is successful in the present environment, to persist it must be adaptable to potentially dramatic changes.

Beyond its present behavior, the potential future behaviors of an organism are also encoded by implicit bets within its genome. That is, organisms are predisposed to certain kinds of variation at the expense of others \cite{kirschner2005plausibility}; mutations of a genome result in a range of accessible changes to behavior and morphology, and that range of accessible variation can itself be selected for. Such \emph{facilitated variation} \cite{gerhart2007theory} operates through {
evolved mechanisms like development \cite{carroll2005endless}, canalization \cite{siegal2002waddington}, and genomic organization \cite{kashtan2007varying}; and other mechanisms could yet evolve in the future. For example, there exist mutations that give a person six functioning fingers on each hand, which are enabled by the developmental programs encoded in the human genome; these kinds of modular high-level mutations are more evolvable than if the design for a new finger required independent re-invention. Such \emph{evolvability} of a genome is also subject to selection, insofar as it enables or hinders a lineage's ability to persist.

A genome's accessible variation implicitly encodes a bet about the future opportunities and situations an organism's lineage may encounter; e.g.\ the insect body plan’s dependence on an exoskeleton limits the size of organisms evolved in a different way than that of animals with an endoskeleton, and different body plans enable different flavors of phenotypic variation (e.g.\ wing size in birds; tail size in lizards). This can be seen somewhat analogously to how ML algorithms themselves develop over time as researchers extend them and apply them to new domains and contexts (e.g.\ the many different variants of LLM activation functions, positional encodings, and variants of attention layers); the difference is that evolution explores such architectures and systems autonomously.

Thus in total, a genome implicitly encodes a hypothesis about the present and possible futures, which can be falsified if it ceases to persist (either through it not reproducing, or through all of its children’s lineages not persisting). As in science, such a hypothesis is never completely validated, as tomorrow's environmental conditions or the adaptation of an organism vying for the same niche may undermine a lineage's prospects. Importantly, there is no fixed time horizon for falsification (in contrast with current RL, as we will later see).


\subsection{Evolution as Open-ended Generator of Novelty}
\begin{quote}

``The beauty of the living world I was trying to photograph was spread before me like a tapestry of evolutionary innovation, with one outlandish strategy after another. A flower shaped like a bucket to trap insects. A fish that farms algae. Another that stabs its prey with its snout. Life evolved to fill every conceivable niche.''

\hfill --- Claude 3.5 Sonnet hallucination attributed to Frans Lanting
\end{quote}

Evolution, like science, is antifragile \cite{taleb2014antifragile}: the system as a whole improves from shocks (e.g. in science, from experiments that empirically refute a hypothesis, or in evolution, from environmental changes that cause extinction \cite{lehman2015extinction}). 
More broadly, {
there are pressures in evolution to diversify and expand through possible behaviors and niches} \cite{stroud2016ecological,gilbert2020natural,stanley2015greatness}, such as to escape competition \cite{pfennig2012evolution,gilbert2020natural} or predation \cite{johnson2020predators}.
Over evolutionary time, this antifragility and pressure to diversify causes evolution to accelerate, which is known as the evolution of \emph{evolvability} \cite{dawkins2019evolution,payne2019causes}. Common critiques of evolutionary algorithms \cite{yampolskiy2018we} tend not to recognize this \emph{self-accelerating antifragility}; while evolution may start slow, like science, it accumulates an expanding repertoire of powerful building blocks (e.g.\ like individual proteins, multicellularity, developmental systems, and neural computation) that can be flexibly recombined to speed up evolutionary search and realize products beyond what humans can currently design (e.g.\ robust human-level intelligence).

An important condition for such acceleration is that opportunities for meaningful adaptation and diversification must not permanently halt \cite{lehman2013evolvability}. This condition highlights a key difference between biological evolution and most ML algorithms (including evolutionary algorithms): that biological evolution is an open-ended search process \cite{stanley2015greatness,soros2017open} with no fixed end point, and tends to creatively and endlessly diverge and innovate}. The tree of life bootstrapped itself into vastness, and individual ecosystems intricately intertwine the interests of diverse organisms; adaptation and diversification of one species provides novel opportunities and challenges for others \cite{pfennig2012evolution}, and so on.

Because an organism's environment is largely shaped by other evolving organisms, and it is impossible for one organism to anticipate the possible adaptations of all others that interact with it, novel situations of KU become unavoidable. Thus evolution is the self-same accelerating engine of generating KU and selecting for it: novel adaptations of one species generate KU for other ones, which tests their robustness. In this way, evolution's open-ended creativity is important for two separate reasons: Continual diversification of creatures continually generates (1) 
diverse implicit bets about how to persist through future and (2) new adaptations that constitute unforeseen situations to filter the bets of other organisms. 

In conclusion, the open-ended and accelerating creative divergence of evolution continually refreshes the set of diverse bets made about strategies for persisting the future. In contrast, ML algorithms rarely include continual and expansive diversify-and-filter strategies for dealing with the unknown future; nothing in principle prevents them from doing so.

\section{Reinforcement Learning's Formalisms Limit Robustness to Knightian Uncertainty}
\label{sec:rllimits}

\begin{quote}
``A picture held us captive. And we could not get outside it, for it lay in our language and language seemed to repeat it to us inexorably.''

\hfill --- Ludwig Wittgenstein
\end{quote}

This section explores the formalisms underlying reinforcement learning (RL) as a representative example of how convenient assumptions implicitly limit ML's engagement with the problem of KU; without loss of generality, similar assumptions may apply also to unsupervised and supervised learning. 

\subsection{RL's Core Formalism: The Markov Decision Process}

The foundational formalism of RL is the Markov decision process (MDP; \cite{sutton2018reinforcement}), wherein an agent in a fixed environment iteratively observes a state and reward, chooses an action in response, and transitions to a new state. The MDP formalism makes many assumptions that are often violated in practice, which has motivated many variations fit to  particular settings. For example, the partially-observable MDP (POMDP; \cite{spaan2012partially}) models  environments where an agent must act without complete information; Markov games \cite{littman1994markov} apply to mixed-incentive multi-agent environments; and bounded-parameter MDPs (BMDPs;\cite{givan2000bounded}) extend MDPs to environments where transition probabilities and rewards are not precisely known. 

That RL as a field continues to diversify its formalisms might seem to undermine the argument that core assumptions limit RL's ability to deal with unknown unknowns. In some sense, RL has no core assumptions, and there is a strong agenda within RL to deal with its shortcomings by relaxing assumptions in different ways to tackle more realistic problem settings. Yet as others have pointed out \cite{abel2024three}, there are implicit dogmas in RL that may sometimes cloud its larger ambitions.
For example, from the perspective of open-world unknown unknowns, nearly all MDP relaxations suffer from the same pathology: They require a researcher to \emph{quantitatively anticipate} important properties about \emph{qualitatively unknown} future risk.

In more detail, the price of relaxing an assumption is that it makes deriving an effective and efficient algorithm more difficult -- especially one benefiting from formal guarantees; thus in practice, assumptions are only partially relaxed into new unrealistic assumptions, often through offloading onto the experimenter the tall challenge of anticipating all possible domain variation. 
For example, BMDPs require an experimenter to explicitly describe uncertainty in the distribution of transitions and rewards, which is a poor fit to describe the qualitative uncertainties inherent in an unknown future \cite{taleb2010black}. More directly, does the functional robustness of the zebra mussel result from its robustness to a certain level of uniform perturbations to its observations? Certainly robustness to input signals is useful; but much more pertinent is the robustness of its general living strategy (e.g.\ how well does its conception of ``threat'' and its response to it generalize to rare but high-stakes situations)?

Notice, in contrast, that biological evolution proceeds without any explicit formalism, of risk or otherwise. Organisms implicitly encode within their genomes how they will deal with risk, and unsuccessful bets are culled through natural selection. This is not so different from how humans and corporations deal with risk over long time-horizons; individuals take different approaches, and whichever works in practice tends to be elevated. An interesting question this raises is whether RL's attachment to a formal basis could at times prove an obstacle to the field's progress, an issue that we return to later.

The next sections examine consequences of RL's formalisms in more detail, highlighting how specific aspects of them contribute to blindness to KU. Because there are too many variations of formalisms and algorithms to address comprehensively, in what follows we assume the RL algorithms most popular in practice, operating under the typical POMDP formulation of RL. We do not intend in what follows to deny the progress and promise of RL algorithms or
undercut RL's many economically and scientifically exciting successes; instead, the hope is to point towards
a subtle emergent issue, which is that many independent small assumptions as a whole paint a world where KU does not exist, or if it does, it does not require direct engagement.

Readers less interested in specifics of the RL formalism can proceed to Section \ref{sec:discussion}, where more general implications of this paper are discussed; what is most critical below is that many assumptions in RL are fitting for a closed world, but definitionally preclude engaging with messy facets of KU in an open one.

\subsection{RL is Time-blind}

We argue that the unfolding of time in an open world is fundamental to engaging with KU. 
However, there are several ways in which RL formalisms can blind RL to time. First, RL often assumes a world frozen in time, i.e.\ that the distribution of training environments is fixed, and that the deployment environment comes from the same distribution and is similarly held in stasis. Secondly, the discount factor in the most common formulation of RL's objective implies that there is a fixed time horizon beyond which an RL agent is blind to its impacts. Third, RL assumes that time cannot leak beyond the boundaries of each discrete episode. Finally, RL most often treats the data it trains upon as timeless; that is, most RL algorithms update their policy by processing a mixed batch of data, unmoored from when in the agent's life they occurred.

\subsubsection{Deployment and training environments are assumed identical and static}

In RL, while generalization is equally as important as in supervised learning, most formalizations make no distinction between the train and deployment environment \cite{cobbe2019quantifying}. As a result, it is not surprising that many RL solutions are brittle \cite{song2019observational,heaven2019deep,huang2017adversarial}, as overfitting is to be expected from training on the test set. Domain randomization, procedural content generation, adversarial training, and other methods can encourage generalization \cite{cobbe2019quantifying,akkaya2019solving,pinto2017robust}, although for KU the higher-order problem then becomes identifying the types of variation representative of novel risks an open-world agent will plausibly face in the future. In other words, the issue of KU is not engaged with by the formalism to be learned by data; instead it is off-loaded onto the experimenter.

The fundamental challenge is that it is hard to characterize the flavor of generalization that is desired from first principles when dealing with out-of-distribution extrapolation in the abstract \cite{kirk2023survey} (without a notion of a larger world evolving in time that generates new situations). Heuristics such as NN simplicity \cite{cobbe2019quantifying}, or disentangled NN representations \cite{higgins2017darla} can perhaps often be  useful \cite{kirk2023survey}, but it is unclear that a perfect such heuristic or combination thereof is universally appropriate. Indeed it seems doubtful that humans have yet arrived upon ideal principles for dealing with open-world KU. Evolution's strategy (see Figure \ref{fig:diversify}) or human ones perhaps can provide inspiration; and perhaps there is a way to frame the problem that brings more of the temporal unfolding of an open world inside the formalism.

\subsubsection{Fixed Time Horizon}

The standard objective function for reinforcement learning is to find a policy that maximizes the expected return $G$, which is a discounted sum across time $t$ of rewards $R_t$:

\begin{equation}
G = \sum_{t=0}^{\infty} \gamma^t R_{t+1} = R_1 + \gamma R_2 + \gamma^2 R_3 + \dots
\end{equation}

Prominent in this equation is the discount factor $\gamma \in [0, 1)$, whereby the importance of a reward or punishment decays across time. With increasing $t$, $\gamma^t$ approaches 0, and the algorithm becomes indifferent to outcomes, implying that (1) credit assignment becomes impossible when actions and consequences are sufficiently distant in time, and (2) equivalently, an RL algorithm is indifferent to catastrophic events beyond its time horizon. As a result, RL is most frequently applied to environments requiring only short time-scales, like games \cite{hung2019optimizing}.

While time horizons can be extended by drawing $\gamma$ closer to 1, larger $\gamma$ can slow convergence of RL  due to increased variance \cite{hung2019optimizing,schulman2016optimizing}, and credit assignment across long time-horizons is significantly challenging in itself \cite{hung2019optimizing,arjona2019rudder}.
While time horizons can also be extended through the use of hierarchical RL (HRL; \cite{barto2003recent}), HRL is rarely adopted in practice for high-profile results; one reason is that HRL induces challenging exploration and optimization problems \cite{gupta2019relay}. 
Even given HRL, or other methods for long-term credit assignment in RL \cite{hung2019optimizing,arjona2019rudder}, it is challenging (and to our knowledge far beyond the state of the art) to scale them to timescales easily managed by evolution, i.e.\ managing risk across many years.
Some RL algorithms maximize average reward, which circumvents these issues, but for theoretical reasons they then often assume ergodicity \cite{mahadevan1996average}, which implies that irrecoverable failures in the domain are impossible; this assumption directly contradicts the aim of robustness to KU (which often involves unrecoverable mistakes). 

Interestingly, the time horizon across which evolution manages risk has no intrinsic limit. An organism can survive for years before reproducing, and actions taken near the beginning of its life can impact whether it reproduces or not. More broadly, a branch on the tree of life can end up being a dead-end, even after persisting for many, many generations, if an implicit assumption it encodes (e.g.\ a qualitative approach to making its living) is invalidated by a change in its environment or the adaptation of another species. For example, the lifestyle of the trilobite was successful for over 200 million years, before it went extinct, driven into decline by predation from better-adapted species such as early jawed fish, which had more efficient feeding mechanisms and locomotion \cite{fortey2010trilobite}.

\subsubsection{Episodes are Hermetic}
\label{sec:episodes}

Relatedly, most RL algorithms break evaluations into distinct episodes, which is a complete sequence of states, actions, and rewards spanning an initial state to a terminal state \cite{sutton2018reinforcement}. For example, a legged robot may navigate from a starting location to a goal location. The assumption is that each such episode is strictly independent from others, and generally, that all episodes are drawn from the same timeless distribution.
However, in the real world, each episode changes the environment in ways both subtle (e.g.\ a robot’s motors will wear differentially from turning left more than right) and obvious (e.g.\ where the robot ends up after the ‘end’ of the episode is where it will ‘begin’ at the next \cite{lu2020reset}); further, in open worlds, the environment is constantly changing, often in response to adaptations of the agent \cite{brooks2024selfdriving,lewandowski2021influence,xu2024llm}. 

In open-world situations, where an agent is deployed in an on-going basis, and interacts with other adaptive agents (e.g.\ humans or other learning agents) in a persistent environment, a formalism that imposes episodic limits is at odds with the agent's actual situation, and is therefore likely to cause fragility. That is, optimizing as if episodic boundaries are real will likely yield behavior over-optimized for episodic boundaries; in effect, internalizing incremental performance gains while externalizing risk and fragility. For example, a robot may complete a task marginally faster if it sacrifices motor wear for increased speed, which may have negligible impact within an episode but compound across the robot's life.

Evolution, in contrast, has no rigid episodes. While the life-cycle of an organism is a natural unit of evolutionary time, past life-cycles impact present and future ones in important ways, unlike the strict boundaries of episodes in RL. That is, a parent organism may help its offspring, the `starting state' of an offspring (e.g.\ where it begins its life, and how much energy it begins life with) is dependent upon its parent, and environmental changes from past organisms can persist across time (e.g.\ a beaver's dam, the composition of soil, or the population dynamics between predator and prey).

\subsubsection{Data is Temporally Undifferentiated}


From the point of view of an e.g.\ off-policy RL algorithm like DQN \cite{mnih2013playing}, within a training batch there is no distinction between data from recent episodes and from far earlier ones. In other words, the algorithm cannot distinguish mistakes made by ``younger'' and ``older'' versions of the agent. For on-policy algorithms, past experience collected from prior policies has limited value. In both cases, neither the RL algorithm nor the policy can explicitly generalize from the trajectory of prior versions of itself's mistakes (e.g.\ a person can make higher-level abstractions about the kinds of mistakes they have previously made years ago, because they can recall those mistakes, and how they later learned to avoid them). Similarly an LLM trained with RLHF does not learn from an explicit history of their past mistakes earlier in training, nor do deployed LLMs reason explicitly from mistakes they made during RLHF. The point is that the abstractions applied in RL training cut off important temporal information that evolution has access to and often does exploit.

RL algorithms can struggle to learn quickly from rare catastrophic events (e.g.\ the rare event may be swamped in gradient updates by more mundane data; whereas a child will not put their hand on the hot stove after encountering it once). To address this point, researchers have explored policies based on episodic memory \cite{blundell2016model,lengyel2007hippocampal}, wherein explicit memories of 
past episodes serve to guide present agent action. However, as of yet they are not common in practice, and the integration of episodic memory into a larger RL framework often requires experimenter-designed heuristics for what memories to place into memory, how to represent them, and when to recall them.

Interestingly, RL agents themselves rarely receive time as an input \cite{pardo2018time}, meaning the agents themselves are fundamentally blind to absolute time (although policy classes like recurrent NNs, or positional encodings in LLMs enable relative time-keeping within an episode), i.e.\ to know how long they have been trained for, or how many episodes they have seen.

In contrast, the RL algorithms embodied by intelligent animals make flexible use of time and history. For example, a wide mix of model-free, model-based, and episodic learning mechanisms have evolved across different families of animals \cite{wynne2020animal}. Because biological evolution is not tied to mathematical gradients, it is not logically constrained by the memory or computational costs of attempting to calculate gradients across the lifetime of an animal, nor by the greater problems of vanishing or exploding gradients that often complicate gradient-based meta-learning algorithms \cite{antoniou2019train,vettoruzzo2024advances}. Of course, there is no free lunch, and evolution cannot transcend the information-theoretic challenges of credit assignment or the energetic costs of computation \cite{kempes2017thermodynamic}, yet it is hard to deny that the learning algorithms evolution has uncovered are surprisingly efficient and robust relative to those in RL. Moreover, on an evolutionary time-scale, there are evolutionary equivalents of the notion of generalization \cite{kounios2016resolving}, and evolution can ``learn'' from rare catastrophic events like extinctions, which can serve to accelerate it \cite{lehman2015extinction}.

\subsection{RL has a Thin Conception of Risk}

While RL's standard objective (expected value) engages with risk, and there exists many approaches that aim to satisfy more nuanced measures of risk \cite{garcia2015comprehensive}, problems can result from optimizing an open-world agent for such metrics under closed-world assumptions. RL can effectively externalize risk outside the bounds of its formalism, similar to arguments in \cite{taleb2010black,taleb2014antifragile} (or Figure \ref{fig:truerisk}), creating a false sense of confidence. 

\subsubsection{Counter-intuitive Implications of Expected Value}

Most RL algorithms explicitly optimize for the policy that maximizes the expected reward of an agent \cite{sutton2018reinforcement}, i.e.\ the reward an agent attains when averaged over independent episodes. While seemingly an intuitive notion of performance, it has counter-intuitive implications and subtleties. For example, expected reward can incentivize risk-taking and fragile optimizations that achieve slight benefits in training environments over broader robustness \cite{mihatsch2002risk}.

In particular, expected reward does not consider the \emph{variance} of rewards \cite{mihatsch2002risk,flageat2024beyond}. That is, in most real-world situations not only the average outcome, but the variance across outcomes is instrumentally important. For example, it is obvious that when deciding between two options, one that leads to a predictable outcome is significantly different from one that has wildly divergent outcomes (e.g.\ if one taxi offered an average arrival time of 40 minutes, plus or minus 5 minutes; and another offered the same average arrival time, plus or minus 20). Yet the most common RL objectives and evaluation metrics are blind to the real qualitative differences underlying a similar average score \cite{flageat2024beyond}. 

\subsubsection{Optimization of Risk Measures can Externalize Risk}


The subfield of safe RL \cite{garcia2015comprehensive} does explore many ways of formalizing decision-making under risk, such as optimizing for the worst-case or the variance of returns, although such risk-sensitive optimization is rarely observed within RL's highest-profile results (as it is a more difficult optimization target). Distributional RL \cite{bellemare2023distributional}, while most often is used to optimize for expected reward, does model the distribution of reward, which is a promising broadening of the RL formalism. However, while RL is making progress on practical RL algorithms for managing risk, such progress obscures two deeper philosophical issues.

The first issue is that such risk-sensitive approaches are limited by the \emph{problem settings they are applied within}. In other words, if a risk-sensitive RL algorithm is applied under the assumption of a static distribution of environments (as is most common), what risk it models will not include those of an open future. That is, the algorithm will take e.g.\ the variation of returns into account, but not with respect to unforeseen situations. The result then is a policy that manages known unknowns, but remains vulnerable (and perhaps more so, due to overconfidence) to unknown unknowns. Indeed, recent theoretical work in RL has argued that such formalisms of decision under risk are insufficient to deal with black swan events, even given a static environment \cite{lee2024black}.

The second issue is that the design of the reward function itself (which serves to quantify how catastrophic an outcome is), the risk criterion (which serves to quantify how to trade-off risk and reward), and the environment distribution (which serves to represent the variety of circumstances an agent will experience) are relegated to the experimenter. That is, while pragmatic and useful tools for human-guided risk management, the challenge of dealing with an unknown future is not actually confronted, but instead handed off to experimenter intuition and iteration. In contrast, evolution empirically tunes risk relative to persistence, with no temporal limit (i.e.\ it is capable of managing geologically-long time horizons).

\subsubsection{Insensitivity to Correlated Risk Outside Formalism}
\label{sec:insensitivity}

The most common RL formulation, of training a single policy in a fixed environment, is so common that we are often blinded to a significant difference between the train and test environment, namely that many copies of the same policy are often deployed into the wild post-training. Now, what was previously an uncorrelated risk (e.g.\ of one policy failing in the training environment) can become a highly-correlated risk (e.g. if hundreds of self-driving cars simultaneously face the same rare weather condition, and all fail on the same day; or if the numerous instances of a globally-deployed LLM are all susceptible to the same jail-break). In other words, optimizing for safety in the single-agent case can obtain it narrowly, while creating unexpected emergent situations at many-agent deployment and externalizing systematic risk. 

In a similar way, if catastrophic risk is correlated \emph{across} episodes (e.g.\ as in Russian roulette), then RL will be insensitive to compounding risks if it is then deployed in situations consisting in effect of many sequential episodes. For example, one challenge in the LLM agents paradigm is that of error compounding \cite{dziri2024faith}, where long chains of independent LLM calls end up reliably failing, due to small errors that compound (i.e.\ risk is correlated across LLM calls), even if many of the independent calls in themselves are fairly reliable. This concern is related to the assumption of episodic independence discussed in section \ref{sec:episodes}.
%

\subsection{Implicit Limiting Assumptions of RL Algorithms}
\label{sec:implicit}

In any field, there are often implicit assumptions that go unsaid, such as what elements of the system form fundamental dependencies and what limitations result from them (e.g.\ relying on deep NNs as function approximators), what kinds of research the community values and believes holds long-term potential (e.g.\ creating a domain-agnostic RL algorithm), or how the field's formalisms and benchmarks relate to the field's stated ambitions (e.g.\ how does learning a fixed policy in simulation relate to general intelligence in an open world)? This section highlights how some of RL's implicit assumptions also undercut confronting the challenge of KU.

\subsubsection{Reliance upon NN generalization to OOD observations}
\label{sec:generalization}

In most current RL algorithms, the policy output by training is a NN that maps observations to actions. The neural network serves as a function approximator, e.g.\ in deep RL, a parameterized NN approximates an optimal policy or Q-function. While the learned NN can accept inputs that span the entire space of possible observations (e.g.\ for a vision-driven policy, the space of all images, or for a language-driven policy, the space of all possible input text), the training data for the network is drawn from only a sliver of that greater space. 

As argued in Figure \ref{fig:diagram}, an unforeseen circumstance will often present an observation to the NN that is out-of-distribution (OOD) relative to training environments \cite{sonar2021invariant}, and NN generalization seems a potentially over-optimistic solution to the issue of KU. 
In RL in particular this assumption is suspect, as rare or novel situations can have dramatic implications for what sequence of actions are thereafter appropriate (e.g.\ in driving policies for cars, in program synthesis, or in games like Go). That is, successful generalization may be highly context-specific.

More broadly, the particular NN architecture and its learning algorithm (e.g.\ gradient descent) are a core dependency of an RL algorithm for generalization, and the implicit assumption is that generalization capabilities of such NNs are appropriate and sufficient to deal with rare or novel situations, and that gradient descent is a sufficient algorithm for the job. However, reliable extrapolation is a difficult and fundamental challenge for any function approximator (including deep learning and foundation models; \cite{chollet2019measure,xu2020neural,kumar2019stabilizing,berglund2023reversal,lazaridou2021mind,haley1992extrapolation}), and out-of-distribution generalization is an under-specified problem. 

Furthermore, dependence on the convenience of well-behaved gradients and on the hardware lottery of GPUs \cite{hooker2021hardware} is an invisible hand that implicitly constrains the neural architectures explored by researchers. That is, while in biology it is common to have many feedback connections from higher layers to lower layers \cite{callaway2004feedforward}, nearly all dominant NN architectures structurally limit recurrent connections (i.e.\ \emph{to a layer feeding back into itself}, rather than to lower layers), due to computational and gradient stability issues. It is clear that the streetlight effect \cite{freedman2010scientific} (i.e.\ searching only where is convenient to look) is not a sound principle of search, and we would not knowingly design our algorithms to fall prey to such a bias; of course, for many complex reasons as a research communities we do not necessarily organize our resources in ways consistent with our domain knowledge \cite{stanley2015greatness}.

In contrast, evolution is not committed to a single mechanism for generalizing to unseen states (e.g.\ it does not depend on action as a mapping from observation by way of a single monolithic NN), and often invents and adapts learning mechanisms to handle OOD situations more robustly (e.g.\ surprise, caution, and fear can indicate when one is in a novel situation and temper one's actions) and generalize better from single examples (e.g.\ robust episodic memory). Evolution, unlike RL algorithms with fixed architectures, can adjust all facets of neural architecture, connectivity, and neural learning algorithm, and often composes many separate learning processes together.



\subsubsection{RL algorithms are developed to be domain-agnostic}

In a related vein, robustness to KU is also limited by the aim within RL to create \emph{general, fixed} RL algorithms, i.e.\ algorithms that apply equally to any possible world expressible within its formalism (e.g.\ a POMDP). While a benefit for practitioners seeking to quickly tackle a new domain, and for algorithm designers to have a convenient theoretical base, the generality of an RL algorithm (e.g.\ PPO or Q-learning) implies that the RL algorithm \emph{itself} cannot adapt as the agent incrementally encounters unforeseen situations, especially ones that trigger shortcomings of the underlying RL algorithm (which will actively be sought out in open-world situations where some agents are adversarial).

That is, research in RL often proceeds by human researchers uncovering a latent failure mode of an RL algorithm, and then patching it in some way. For example, the discovery that deep Q-learning suffers from over-optimism motivated double Q-learning \cite{hasselt2010double}; or that gradient updates to a policy may be too large or small, motivating the use of a trust region \cite{engstrom2020implementation}. Yet in an open world, foreseeing all such failures a priori may be implausible. While one could argue that algorithmic generality respects the bitter lesson \cite{sutton2018reinforcement} because it relies on data gathered from the world rather than hand-coded assumptions, the problem is that it is not data-driven enough: The \emph{RL algorithm itself} is a critical bias to search that does not update with data or compute, and cannot learn from its failures in unforeseen situations. 

Fundamental shortcomings built into the algorithm (e.g.\ the POMDP formalism; the implied time horizon beyond which it is indifferent; higher-order patterns in learning failures, like catastrophic forgetting or gradient spikes) remain limiting, and the RL algorithm can benefit only from whatever set of hand-programmed mechanisms were included within it \cite{hessel2018rainbow}. For example, the performance profile of different RL algorithms often diverges across even relatively-similar tasks (e.g.\ different Atari games; \cite{such2017deep,ye2021mastering}), meaning that the different heuristics, assumptions, and implementation details underlying such algorithms hold varying benefit in different contexts; yet in an open world, what qualitatively new contexts will be encountered is unknown, leaving the system designer in a difficult position. 

In contrast, biological reinforcement learning algorithms embody rich priors about agent-specific risks and the qualitative world they occupy, themselves learned from evolutionary experience (meaning that across evolutionary time they have survived encounters with unknown unknowns and adversaries motivated to exploit shortcomings in their adaptations). 

The overall conclusion is that many independent assumptions in RL's formalism, and implicit in the field more generally, contribute to clouding the importance to general intelligence of robustness to an unknown open-ended future.

\section{Discussion}
\label{sec:discussion}

This paper highlights the importance of robustness to unknown unknowns for general intelligence, how biological evolution successfully wrangles with KU, and the intriguing blind spot that ML has for it. This section discusses the implications of these ideas. First, its implications for foundation models, RLHF, and AI safety, then what outside approaches and fields are naturally most promising for confronting the challenge of KU. Finally, the paper concludes with possibilities for synthesizing KU into RL's formalisms.

\subsection{Implications for Foundation Models, RLHF, and AI Safety}
\label{sec:rlhf}

A possible tension in the argument about blind spots in RL's formalisms comes from the obvious success of foundation models, included those fine-tuned using RL through human feedback (RLHF; \cite{christiano2017deep,rafailov2024direct}). If KU is so important, why are foundation models useful across many open-world situations? We believe ultimately there is no contradiction, in that such models are indeed immensely useful, and as argued in the introduction, they still struggle with robustness in intriguing ways \cite{stewart2024surprisingly,zhao-etal-2024-improving,chen2024can,huang2023benchmarking,wang2024can,xie2024osworld,jimenez2023swe,siddiqui2022tesla,npr2024_tesla_probe,herger2024waymo,liu2024curse,dhuliawala2023chain}. More broadly, there has been much concern with the safety of large models \cite{amodei2016concrete,hendrycks2021unsolved,nick2014superintelligence}; many of such concerns relate to issues of robustness \cite{amodei2016concrete,hendrycks2021unsolved}, but KU is rarely mentioned in such contexts. This section thus discusses what KU implies for foundation models in general, RLHF algorithms in particular, and ends with reflection with implications for AI safety.

\subsubsection{KU and Foundation Models}

One powerful role of AI is as empirical philosophy, providing direct evidence towards settling deep questions in linguistics and cognitive science that were previously argued only rhetorically or from our current understanding of human intelligence \cite{mcgregor2023chatgpt,floridi2023ai,piantadosi2023modern}, e.g.\ can intelligent behavior be learned through symbols alone, without grounding in experience? In short, AI broadens our understanding of intelligence by creating divergent implementations of it \cite{lehman2015investigating}. ML grants the freedom to isolate particular abstractions of intelligence into algorithms, and explore how they compare and contrast with human intelligence, thereby helping us to better understand the broader space of possible intelligent agents. We learn, for example, that what comes easily in human development is difficult for some types of machines, or vice versa (e.g.\ Moravec's paradox \cite{moravec1988mind}). By leaving KU out of our formalisms, we thus learn how far we can nonetheless get, which is interesting in its own right; the space of failures left after scaling methods blind to KU may then reveal, through negation, what robustness to KU ultimately contributes to general intelligence.

From this point of view, one interpretation of the success of RLHF and current LLMs more generally is that an extremely broad distribution of training data (e.g.\ the internet and large collections of human feedback), coupled with the surprising generalization capabilities of increasingly large NNs, does indeed capture a large range of adaptive (and impressive) behavior -- even when the training formalism mismatches the open-world nature of the deployment setting. In other words, seemingly much of language-based behavior can be captured by a form of interpolation in a highly-abstracted space, given sufficient data. This is no doubt an incredible scientific and philosophical discovery, and one of great economic importance. For many business-as-usual situations, the ability to deal with KU
may not be necessary for adequate behavior, given sufficient coverage of similar situations in training.

Conversely (and as argued by others \cite{chollet2019measure,mitchell2021ai}), the failures of large models \cite{stewart2024surprisingly,zhao-etal-2024-improving,chen2024can,huang2023benchmarking,wang2024can,xie2024osworld,jimenez2023swe,siddiqui2022tesla,npr2024_tesla_probe,herger2024waymo,liu2024curse,dhuliawala2023chain}
have interesting implications as well. For example, the ARC prize has (until recently) stymied LLMs despite the seeming simplicity of its task \cite{chollet2024arc}; its design was explicitly motivated by the theory that ML models struggle to acquire skills that have little support within their training data \cite{chollet2019measure}. This theory  fits well with the narrative in this paper around KU, as a qualitatively new task with little support in the training set is effectively an unknown unknown from the perspective of the model. 
 Yet of course, general intelligence must include the ability to acquire new skills on the fly; and many other intriguing failure patterns of LLMs can be related to KU-blindness, such as the continued fragility of AI agents (see Section \ref{sec:insensitivity}).

In conclusion, the arguments in this paper apply also to foundation models, which do remain interestingly fragile, especially to unforeseen situations; we should expect so if the formalisms motivating both their pretraining (supervised learning) and post-training (reinforcement learning) exclude by definition the possibility of worlds with unknown unknowns. The exciting possibility is that through revising such formalisms (discussed later) or otherwise bringing KU within the realm of ML, it might be possible to train large models that are much more robust to the open world.

\subsubsection{KU and Reinforcement Learning through Human Feedback}

Because RLHF is among the most impactful current RL algorithms, this section explores the implications of KU for it in particular. In the most common RLHF setups, a \emph{reward model} that represents human preferences is trained from preferences collected across a representative set of tasks (the \emph{task distribution}). This reward model then acts as RL's reward function, enabling RL to fine-tune a foundation model to produce more-preferred responses for all the tasks in the task distribution.

RLHF has been highly successful in practice, and yet the arguments detailed so far also apply to it. For example, the task distribution used in RLHF is an important and messy consideration, and if qualitatively novel tasks emerge after a model is deployed, the ability of a model to gracefully handle such tasks depends on its robustness to KU; this issue may underlie e.g.\ fragility to novel jailbreak attacks and underperformance on tasks like ARC that are designed to represent new but learnable skills. Given the breadth of the world, it seems intractable to anticipate a priori the complete space of qualitative tasks a widely-used model may later be asked to tackle.


Interestingly, KU may also impact the RLHF training process itself, because RLHF depends upon a model of human preferences. In particular, the KU perspective suggests a  tension between the creativity we might want from RLHF (i.e.\ for learning to uncover qualitatively novel superhuman solutions to given tasks), and the ability for the reward model to successfully \emph{recognize} such creativity. That is, a truly novel strategy for solving a problem may be qualitatively out-of-distribution for the reward model (which is trained to differentiate better and worse responses under closed-world assumptions). If a highly-creative output effectively becomes KU for the reward model, then to realize more of RLHF's potential may require addressing KU in their design. An intriguing empirical study to investigate this hypothesis is to extend a reward model benchmark suite \cite{liu2024rm,lambert2024rewardbench}, and measure if reward models struggle to recognize divergently creative (but objectively correct) solutions. Note that these concerns apply also to methods like RL from AI feedback \cite{lee2023rlaif} and constitutional AI \cite{bai2022constitutional}, wherein foundation models themselves act as reward models for an RL process.

Another interesting consideration is that while the RL formalism underlying RLHF does not directly incentivize wrangling with KU (as per earlier arguments), the training data of LLMs does include semantic high-level information \emph{about} KU itself and the challenges of navigating an uncertain future (e.g.\ Knight's work on KU is likely in common training corpuses, as will be this paper). Thus in theory, models may internalize the problem of KU on some level, and perhaps can be more attuned to KU if so prompted (an idea which is worth future study). Yet it remains an open empirical and logical question to what extent training a model on the \emph{semantic information} necessary to transcend its training paradigm can indeed remedy such limitations; one point of optimism is that humans are able to sometimes understand and transcend their evolutionary biases.

Additionally, LLM training data includes within it transcripts of human reasoning, and reasoning indeed is part of the process through which humans deal with KU. Thus models with reliable and general reasoning may help in some ways with KU \cite{chollet2024o3}. However, for the many reasons articulated earlier in this paper, we believe that reasoning by itself is insufficient for addressing KU; for one, reasoning is a general tool that can easily enable operating in ignorance of KU as much as grappling with it  -- it deeply matters the nuance with which reasoning is applied \cite{kay2020radical}. Secondly, if LLM reasoning itself is simplistically integrated into the language model paradigm (i.e.\ with more tokens relating to reasoning added to training corpuses), an open question then becomes how robust to unforeseen reasoning scenarios it will be (i.e.\ is it indeed general reasoning?).

\subsubsection{KU and AI Safety}

Intuitively, robustness is of particular interest in safety-critical domains (e.g.\ self-driving cars, mental health counseling, medical advice and analysis, etc.), and in adversarial situations in which creative, unforeseen attacks are incentivized (e.g.\ as in LLM jail-breaking, or exploits of self-driving cars). More deeply, robustness to distributional shift between training and deployment is a general concern within AI safety \cite{amodei2016concrete,hendrycks2021unsolved}, and we thus argue that KU has research implications for that community.

Perhaps unsurprisingly, one conclusion is that current framings of robustness to distributional shift could benefit from more direct confrontation with unknown unknowns; in particular, through (1) more fully acknowledging the creative process through which situations of KU emerge in the real world, (2) giving up on the possibility of fully anticipating and training upon all possible domain shifts, and instead (3) encouraging the ability to recognize situations of KU, and to adapt dynamically to them at test-time.

That is, in open-world situations, qualitatively novel situations and improbable edge-cases emerge naturally across time, often due to the creativity and adaptability of other intelligent agents (the billions of humans on Earth, markets, governments, and other institutions). As alluded to above, for a foundation model deployed through an API, it seems plausibly intractable to anticipate the open-ended decentralized and collaborative creativity of those seeking to leverage LLMs for new applications, or to elicit interesting failures from them, or jailbreak them to cause misuse. Furthermore, the outputs of LLMs themselves impact the digital world (e.g.\ the rise of LLM-generated webpages, enable new apps that LLMs feasibly will interact with), which other actors in the world accordingly adjust to as well. In other words, the real world is in a constant state of bubbling creative distribution shift.

In acknowledging that such models will inexorably encounter many unforeseeable situations and edge-cases when deployed in an open world, the consequence is that it may be necessary to find approaches to handle KU other than the dominant strategy of \emph{anticipate-and-train}. That is, the dominant approach for robustness in large real-world models is simply to accumulate as much diverse and relevant data as possible (i.e.\ anticipate), and then to train models upon them, in hopes that NN generalization will be sufficient to handle KU; an extension is to incrementally patch models through human or automated red-teaming \cite{samvelyan2024rainbow}, in effect, another round of anticipate and train. Methods developed specifically to encourage robustness to distributional shift similarly attempt to create many environmental variants (as in domain randomization \cite{tobin2017domain} or adversarial robustness \cite{pinto2017robust}) and train upon them. As described in Section \ref{sec:openworld}, shifting to the meta-learning paradigm does not in practice necessitate moving beyond this anticipate and train paradigm.

Such existing methods are practical and useful, and the critique is not that we should not attempt to use our current tools to help models be as robust as possible. Rather, the overall problem with such approaches is that there is a difference between (1) training in parallel upon anticipated situations, practicing upon them repeatedly to learn an optimal policy that is then deployed, and (2) adapting sequentially across a lifetime of qualitatively novel challenges in a way that improves one's core ability to recognize and work with situations of true uncertainty. The claim is that there is a hard-to-pin-down hole in the current way we formalize our methods
that we hope this paper helps bring to light.

Finally, to end on a broader note, as researchers in ML we ourselves lie in a position of Knightian uncertainty with respect to the development of AGI or superintelligence, what their nature would be if developed, and their safety implications for society \cite{sunstein2023knightian}. That is, the unfurling of science is unpredictable, futurology is littered with 
failed predictions, and the injection into society of a strange and powerful new technology is guaranteed to have unforeseen consequences, both bad and good. This is not to discount our laudable efforts to anticipate the risks to come \cite{hendrycks2021unsolved,nick2014superintelligence,amodei2016concrete,hanson2016age,bommasani2021opportunities}, but instead to suggest that we may need to yet broaden them. If the past is any guide, we should embrace that whatever our beliefs about how close or far we are from cracking intelligence, or what a machine agent much smarter than us would be like in practice, that the realities are likely to greatly surprise us.

\subsection{The Promise of Artificial Life}

One way of interpreting the argument in this paper is that it advocates for evolutionary algorithms \cite{de2017evolutionary} over other ML methods, in analogy with biological evolution's robustness in dealing with KU; we do not intend to take this position. Biological evolution is quite distinct from traditional evolutionary algorithms (EAs), which most often abstract evolution as a black box optimization (BBO) process \cite{lehman2010revising}. Indeed, from the lens of KU, the setting of BBO could be critiqued in much the same way as RL, because BBO is often formalized as global optimization, i.e.\ the search for an optimal policy in a fixed environment; and while EAs are population-based, and thus are capable of exploring diverse policies simultaneously, in practice, in most EAs the population quickly converges \cite{lehman2011abandoning} due to global competition among all individuals. In other words, most EAs would not satisfy the four factors we hypothesized  are responsible for biological evolution's robustness to KU (see Figure \ref{fig:diagram}).

However, there are many ways to abstract evolution into an algorithm \cite{stanley2015greatness,lehman2010revising}. That is, many processes optimize, but few continually produce interesting novelty for a billion years, weaving together diverse, adaptable solutions into complex ecosystems. In contrast to EAs for optimization, the field of artificial life (ALife; \cite{langton1997artificial,bedau2007artificial,adami1998introduction}), especially within the open-ended evolution community \cite{packard2019overview,taylor2016open,soros2017open}, attempts to create evolutionary processes more directly inspired by biological evolution's tendency towards ongoing open-ended creativity.

Simulated ALife worlds, such as Polyworld \cite{thearling1994evolving}, EvoSphere \cite{miconi2008evosphere}, or Avida \cite{ofria2004avida} instantiate 2-D or 3-D simulations where creatures evolve to compete for resources. The hope is that it is possible to set up the initial conditions (often with a seed organism capable of replication) in a rich simulated world such that the resulting dynamics of competition and evolution produce an ongoing proliferation of diverse innovation in complex ecologies, as happened on Earth. If successful, such systems should produce more KU-aware solutions, as they would satisfy the criteria hypothesized here. 

However, it remains an unsolved challenge how to engineer such a rich world wherein running the system results in long-running open-ended evolution \cite{soros2017open,bedau2000open}. Yet in comparison to ML, relatively few resources have been expended in pursuit of ALife, especially when juxtaposed with the intellectual grandeur of what it represents: The understanding of the principles for universe-design that satisfy the necessary conditions for continual and unbounded creativity \cite{soros2014identifying,soros2017open}. Given increases in the scale of possible computation, and the possibilities for making inroads into core aspects of intelligence that RL may be skirting (i.e.\ robustness to KU), we believe this field of research is notably underappreciated; intriguing recent work has explored the possibility for foundation models to help accelerate ALife experimentation \cite{kumar2024automating,nisioti2024text}, and much more synthesis between ML and ALife may be possible \cite{nisioti2024text}. An interesting possibility is that ALife may benefit from scale in similar ways that foundation models have; more can be qualitatively different.

While recent work has demonstrated that ALife populations can adapt to sudden changes in domain \cite{hodjat2024domain}, to our knowledge simulated ALife worlds have not been investigated for their potential to produce robustness to the open unknown. Future work can attempt to engineer ALife worlds that encourage robustness to KU, perhaps informed by the four conditions proposed in this paper (Figure \ref{fig:diagram}). An interesting if speculative question, is if evolved ALife architectures and learning rules, potentially quite alien from ML architectures, could handle KU better than a general-purpose deep RL algorithm? And if so, could lessons be learned from them to inspire new ones?

\subsection{The Promise of Open-endedness}

Beyond creating artificial worlds, the field of open-endedness \cite{soros2017open,hughes2024open} attempts to abstract the \emph{engineering} principles for ongoing creative search in a way that is domain-independent, similar to the way that human open-ended search can be generalized to almost any design space, e.g.\ to create on-going novelty in  architecture, engineering, songs, beverages, scientific papers, artwork, algorithms and inventions. For example, the POET algorithm attempts to continually create new problems for itself to solve \cite{wang2019paired}, in a similar way to how the ML research community challenges itself with new benchmark tasks. The intuition is that successful open-ended search is highly-related to KU, as it will continually create qualitative novelty (as happens in e.g.\ science or art); in the POET example, a new benchmark may provide a unforeseen challenge for the solver.

A possible benefit of this kind of domain-independent approach over ALife worlds is that open-endedness more directly  relates to real-world problems. That is, even if open-ended evolution in ALife is successful, how to make practical use of the agents evolved in simulated worlds may not be trivial \cite{ecoffet2020open}. Further, open-endedness can  leverage recent advances in ML (such as LLMs) to circumvent the need to evolve intelligence from scratch \cite{lehman2023evolution,zhang2023omni}, which could make such an approach more computationally feasible.

Indeed, there is some evidence that open-endedness can help with the problem of generalizing out-of-distribution \cite{dennis2020emergent,samvelyan2023maestro} in specific RL tasks, albeit in limited ways that do not include new qualitative dimensions. We believe that generalization to new qualitative dimensions may require open-ended invention of new and specialized learning algorithms and architectures for agents, similar to the spirit of AI-GAs \cite{clune2019ai}; in other words, there is no reason to expect that the out-of-the-box generalization capabilities of deep learning architectures should be optimal for unknown unknowns. As described earlier, biological organisms have many special adaptations that let them generalize and fail gracefully (e.g.\ a generalized fear of the unknown, specific learning mechanisms for near-catastrophes and painful situations); we could leverage open-endedness techniques to continually invent such mechanisms specifically fitted to the affordances of particular agents in a particular domain. Hybrid-evolutionary methods like population-based training \cite{jaderberg2017population} are a promising step towards allowing aspects of RL algorithms themselves to adapt to their circumstances, and recent work has learned agent architectures through an LLM-based open-endedness method \cite{hu2024automated}; perhaps such techniques could be adapted to target the problem of robustness to unknown unknowns.

One challenge for open-endedness research with respect to KU is that most
applications of open-endedness involve single-agent or few-agent learning in a relatively simple episodic world across short time horizons. Further, the highest-profile open-endedness results generally leverage out-of-the-box RL algorithms to train agents, and thus we believe are subject to many of the criticisms described so far with respect to KU, even if embedded within a larger open-endedness loop. What may still be lacking in open-endedness research is a deep synthesis between simulated ALife worlds (of many diverse agents co-evolving in a persistent complex world) and that of the abstracted open-endedness environments (of single agents solving tasks within a domain of real-world relevance).
For example, games such as minecraft in theory offer open-ended possibilities within a fairly-rich world capable of hosting many agents \cite{grbic2021evocraft}. An intriguing open question is if there exists an interpolative synthesis somewhere on the continuum between ALife and domain-independent open-endedness that enables practical and fruitful engagement with the problem of KU.

\subsection{Revising the RL Formalism}

While it is surprisingly difficult conceptual work to invent formalisms that enfold
previous blind spots, RL has a rich history of such innovations: inverse RL \cite{ng2000algorithms} seeks to identify the implicit objective of
an agent; unsupervised environment design \cite{dennis2020emergent} poses the optimization of challenging environments as part of a larger RL problem that helps single agents to better learn and generalize; distributional RL \cite{bellemare2023distributional} models the distribution of reward rather than simply the average; and inverse reward design \cite{hadfield2017inverse} attempts to bring the human design process of the RL reward itself into the RL formalism.
There is also a history within RL of critique of potential dogmas \cite{abel2024three} and bold hypotheses about what directions will or will not propel RL forward towards its grandest goals \cite{sutton2019bitter,clune2019ai,silver2021reward,vamplew2022scalar}. Thus while this paper largely provides a critique of current RL from the lens of unknown unknowns, one hope is that it can spur new ways of viewing RL and new RL algorithms that better cope with KU. Here we provide suggestions of future work towards that end, starting from more immediate to the more theoretical.

While it may not attack the core of KU, one immediate possibility is to leverage advances in foundation models to generate \emph{qualitative} variations of RL training environments, for direct training, meta-learning, or post-hoc evaluation. That is, one critique of RL is that it substitutes robustness to quantitative unknowns (e.g.\ noisy transitions, observations, or actions; or to quantifications of risk like value-at-risk) for the qualitative unknown of the future. If LLMs have greater understanding of qualitative dimensions in which environments may realistically vary, they can likely be applied to brainstorm a range of qualitative variations (e.g.\ rare but realistic situations) and then implement them by editing the code of a training environment; some work has begun to explore this, although more from the lens of extending agent capabilities \cite{faldor2024omni} than encouraging robustness to KU. Further promise is shown by work highlighting the benefits of qualitative priors from LLMs to aid policy robustness \cite{ma2024dreureka,samvelyan2024rainbow}. One important question would be how to best leverage such additional scenarios to encourage robustness to further unanticipated variation; training directly on imagined scenarios may encourage greater robustness, but still relies on the anticipate-and-train paradigm.

In this way, a complementary possibility is to synthesize what enabled biological evolution's robustness into RL methods. For example, one aspect of KU is to continually accumulate a wide diversity of bets on qualitative futures, which can then be culled by novel situations. One way to realize this in RL would be to apply a diversity-seeking method such as SMERL \cite{kumar2020one} to find many solutions to ``business-as-usual'' training environments, which could be iteratively culled on outlier test-cases that are hand-curated (or are generated by LLMs, as described above). Another synthesis is to explore many-agent forms of unsupervised environment design \cite{samvelyan2023maestro,wang2019paired}, where the system and agents continually provide novel challenges for each other, and adapt them towards the diversify-and-filter paradigm (from Figure \ref{fig:diversify}).

Finally, there is the intriguing challenge of how to formalize KU and robustness to it within the paradigm of RL. The argument here is that unknown unknowns are not well-captured by the current range of extensions to the MDP, and in fact, may be very difficult to capture mathematically at all, as they seem entangled with the nebulous nature of how a complex open world unfolds into the future. In some ways, formalizing KU seems paradoxical: If KU is a valid concept, how can one meaningfully measure the robustness of a policy to what cannot be fully anticipated? Yet, perhaps there are incremental ways to capture at least some of it. For example, incorporating non-stationarity into foundation model pretraining by ordering data relative to \emph{when} it was generated, could grant an environment for competing LLMs to diverify and be filtered over time, potentially encouraging their robustness to culturally new tasks.  
However, ultimately we believe new insights are needed. Perhaps the Lindy effect \cite{taleb2014antifragile} could be formalized to good effect; or a new convincing and practical formalization of what makes a situation qualitatively distinct from others could be derived; or meta-learning could be reposed such that it better captures the qualitative sense of learning how to learn, which includes graceful handling of unforeseen situations at inference-time.

For a moment, however, assume that KU cannot be formalized. The challenge of KU might then be a blind spot of ML precisely because ML tends to organize its sub-fields by anchoring them on distinct formal problem statements; and progress might result counter-intuitively from lessening our reliance upon them (a phenomenon arguably already unfolding in the era of LLMs). While fitting and efficient for closed world problems, this sociological bias of the ML community may not help in tackling the nebulous open-ended world \cite{stanley2015greatness,porter2020trust,kay2020radical}. That is, if evolution has produced solutions much more robust to KU than ML, and has done so \emph{without any formalisms}, obviously they are not logically requisite to its achievement. In this way, it is conceivable that a critical facet of intelligence (robustness to the qualitatively unknown future) potentially lies beyond precise formalization. While such an idea could seem unlikely, or intellectually or aesthetically dissatisfying, it at least bears consideration: The assumption that formalization must lie at the core of ML is itself unproven, and may never be so.
We pose this gauntlet for the RL community: Whether or not KU can be mathematically formalized in a productive way, how to handle it is a challenging and important question; we look forward to what philosophical, algorithmic, or practical advances may result from disproving our thesis, or from attempts to address it.

\section{Conclusion}

This paper highlights how the concept of Knightian uncertainty is a necessary component of general intelligence, how mechanisms from biological evolution contend with it, and how it challenges popular
formalisms in ML. The conclusion is that the ability to deal with unknown unknowns is core to some of machine learning's most ambitious goals, and that truly impressive recent advances in ML may yet skirt the challenge of KU, thus providing a possible explanation for why machine intelligence at times still seems fragile relative to the biological. While a negative result, the exciting consequence is that progress in algorithms and our understanding of intelligence may lie on the other side of direct confrontation with this intriguing problem.

\bibliographystyle{ACM-Reference-Format}
\bibliography{sample-base,ku}

\end{document}